\documentclass[sn-basic,iicol]{sn-jnl}

\usepackage{graphicx}%
\usepackage{multirow}%
\usepackage{amsmath,amssymb,amsfonts}%
\usepackage{amsthm}%
\usepackage{mathrsfs}%
\usepackage[title]{appendix}%
\usepackage{xcolor}%
\usepackage{textcomp}%
\usepackage{manyfoot}%
\usepackage{booktabs}%
\usepackage{algorithm}%
\usepackage{algorithmicx}%
\usepackage{algpseudocode}%
\usepackage{listings}%

\begin{document}

\title[Article Title]{NU-AIR - A Neuromorphic Urban Aerial Dataset for Detection and Localization of Pedestrians and Vehicles}


\author[1,2]{\fnm{Craig} \sur{Iaboni}}\email{csi3@njit.edu}

\author[2,3]{\fnm{Thomas} \sur{Kelly}}\email{tjk39@njit.edu}

\author*[1,2]{\fnm{Pramod} \sur{Abichandani}}\email{pva23@njit.edu}

\affil*[1]{\orgdiv{Department of Computer Science}, \orgname{New Jersey Institute of Technology}, \orgaddress{\street{323 Martin Luther King Jr. Blvd}, \city{Newark}, \postcode{07102}, \state{NJ}, \country{USA}}}

\affil[2]{\orgdiv{Electrical and Computer Engineering Department}, \orgname{New Jersey Institute of Technology}, \orgaddress{\street{323 Martin Luther King Jr. Blvd}, \city{Newark}, \postcode{07102}, \state{NJ}, \country{USA}}}

\affil[3]{\orgdiv{School of Applied Engineering and Technology (SAET)}, \orgname{New Jersey Institute of Technology}, \orgaddress{\street{323 Martin Luther King Jr. Blvd}, \city{Newark}, \postcode{07102}, \state{NJ}, \country{USA}}}

\abstract{
This paper presents an open-source aerial neuromorphic dataset that captures pedestrians and vehicles moving in an urban environment. The dataset, titled NU-AIR, features over 70 minutes of event footage acquired with a 640 $\times$ 480 resolution neuromorphic sensor mounted on a quadrotor operating in an urban environment. Crowds of pedestrians, different types of vehicles, and street scenes featuring busy urban environments are captured at different elevations and illumination conditions. Manual bounding box annotations of vehicles and pedestrians contained in the recordings are provided at a frequency of 30 Hz, yielding more than 93,000 labels in total. A baseline evaluation for this dataset was performed using three Spiking Neural Networks (SNNs) and ten Deep Neural Networks (DNNs). All data and Python code to voxelize the data and subsequently train SNNs/DNNs has been open-sourced.}


\maketitle

\section{Introduction}\label{intro}

\begin{figure*}[t!]
    \centering
    \includegraphics[scale = 0.64]{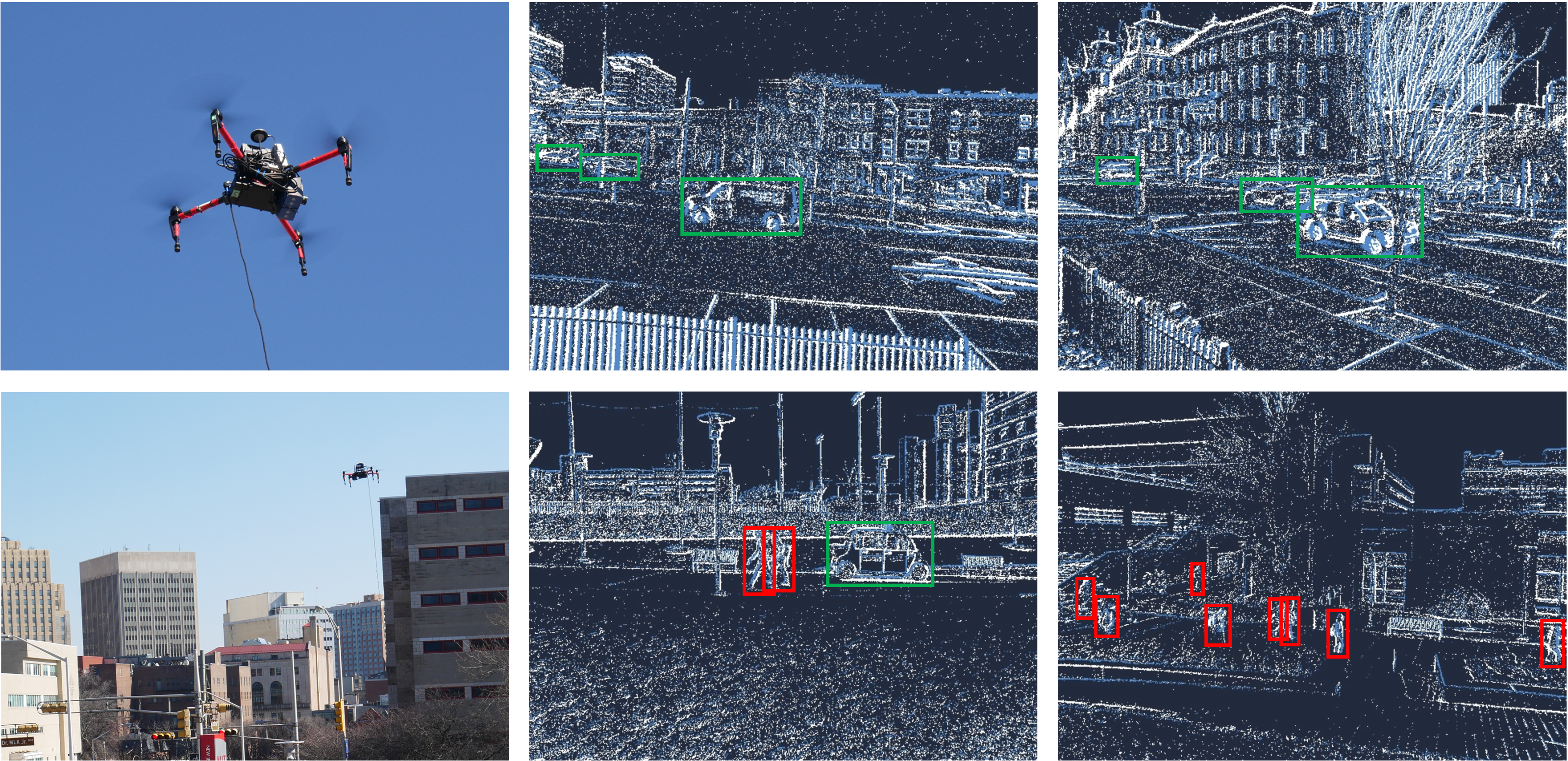}
    \caption{(Left Top and Bottom) The DJI M-100 quadrotor was used to record pedestrians and vehicles in urban environments. A safety rope was affixed to the quadrotor during all flight operations. The recording quadrotor flew at varying heights above a city intersection (Top Mid and Right), a walking path (Bottom Mid), and a campus center (Bottom Right). Bounding boxes created by manual annotators were drawn over frames.}
    \label{fig:event-images}
\end{figure*}

Traditional cameras operate by sampling the field of view at fixed intervals, producing a series of discrete frames that represent static snapshots of the environment. Event cameras differ from traditional ones in that they record asynchronous events rather than continuous frames \cite{gallego2020event}. An event is a change in pixel illumination intensity and consists of the pixel's position $(x,y)$, the timestamp $t$ of the illumination change, and the polarity $p$ indicating that the illumination intensity increased (p = 1) or decreased (p = 0). Event cameras have higher temporal resolution, dynamic range and power efficiency compared to traditional frame-based cameras, making them popular for applications in autonomous vehicles, and robotics/IoT systems where low latency and robustness to lighting conditions important \cite{rebecq2018reconstruction}.

The global aerial imaging market size is projected to grow from \$9.23 billion in 2024 to \$36.70 billion in 2029 at a CAGR of 25.44\%\footnote{https://bit.ly/3TE4Io2}. Large-scale aerial datasets are crucial for developing robust and effective computer vision algorithms. Event cameras are a relatively recent entrant in computer vision compared to RGB cameras. As such, the availability of neuromorphic datasets is relatively low \cite{poker-dvs, sl-animals-dvs, pred18, iaboni2021event, iaboni2022event, postures-dvs, Amir2017}, and only a handful of them feature urban settings \cite{scenes-dvs, miao2019neuromorphic, sironi2018hats, de2020large, perot2020learning}. 
Event cameras, offering solutions to issues like motion blur and dynamic range not addressed by traditional RGB cameras, add a significant capability to the rapidly growing multi-modal aerial imagery industry \cite{aerial-image-market}. The lack of neuromorphic urban datasets featuring aerial footage represents a critical gap, particularly in sectors like search and rescue, target tracking, and infrastructure surveying. Developing a dataset combining event camera data with aerial perspectives is thus beneficial for advancing real-time analytics and operational capabilities in various applications.

As depicted in Figure \ref{fig:event-images}, this study addresses this gap by introducing a neuromorphic dataset consisting of event-recording segments captured from a quadrotor-mounted event camera observing an urban environment during both daytime and nighttime settings. The dataset is fully annotated, enabling the development and evaluation of new event-based vision algorithms for urban settings. 

The main contribution of this study is the open source Neuromorphic Urban Aerial (NU-Air) dataset that aims to advance the study of neuromorphic vision algorithms design. The dataset is uniquely comprised of 70.75 minutes of quadrotor-mounted event camera recordings, segmented into 283 clips of 15 seconds each. These recordings span diverse environment variations including a university campus, traffic intersections, and walking paths in both daytime and nighttime settings. The NU-AIR Dataset has been meticulously annotated for two crucial classes of detection, namely persons and vehicles, with a total of 93,204 annotations in the form of upright rectangular bounding boxes and corresponding class labels. Figure \ref{fig:event-images} illustrates examples from this dataset, along with the quadrotor platform used in the data collection.

Beyond the creation of this dataset, this work also includes a comprehensive ablation study of existing neuromorphic vision algorithms. Specifically, Visual Geometry Group (VGG) \cite{simonyan2014very}, DenseNet \cite{huang2017densely}, and MobileNet \cite{howard2017mobilenets} SNNs has been trained using the NU-AIR Dataset and their performances are compared to previous benchmarks on neuromorphic data.

Additionally, the performance of 10 different Deep Neural Networks (DNNs) is evaluated after training them on the NU-AIR Dataset, further broadening the assessment of neuromorphic vision algorithms. The mean average precision (mAP) is reported for each experiment. The dataset as well as associated Python code to voxelize the dataset and train the SNNs and DNNs is made publicly available through the following repository link: 
\url{https://bit.ly/nuair-data}

The rest of the paper is organized as follows. Section 2 presents related works on open-source neuromorphic datasets. Section 3 discusses in detail the dataset collection setup, annotation process, and relevant statistics. Section 4 presents the baseline experiments and related ablation studies. Section 5 discusses the limitations and key considerations. Section 6 presents the conclusion.

\section{Related Works}
A number of datasets in the literature have reported on pedestrian and vehicle detection in urban environments. The NU-AIR dataset adds to the growing body of open-source datasets across applications such as surveillance systems, autonomous vehicular guidance, and ecological monitoring. \cite{hirano2006industry, mittal2020deep, ramachandran2021review, zhilenkov2018system, sandino2020uav, reckling2021efficient, dilshad2020applications, sandino2021drone, varghese2017power, fromm2019automated, hossain2019deep}.

\subsection{Frame-based Urban Datasets}
Traditional frame-based cameras for object detection of real-world urban scenes have been used to create several pedestrian detection datasets for training and evaluation purposes, including ETH \cite{ess2008mobile}, INRIA \cite{dalal2005histograms}, PRW \cite{zheng2017person}, TUD-Brussels \cite{wojek2009multi}, and Daimler \cite{enzweiler2008monocular}. Larger and more diverse collections such as KITTI \cite{geiger2013vision}, Caltech-USA \cite{dollar2009pedestrian}, UA-DETRAC \cite{wen2020ua}, CityPersons \cite{zhang2017citypersons}, EuroCity Persons \cite{braun2019eurocity}, TRANCOS \cite{guerrero2015extremely}, and TJU-DHD \cite{pang2020tju} have allowed researchers and practitioners to push the limits of algorithmic performance at scale. These sets were recorded by driving or walking through urban areas and provide annotated frames from video sequences.  

Drone-mounted RGB cameras have been utilized to generate multiple real-world aerial datasets. The Urban Drone Dataset (UDD) \cite{chen2018large} features diverse urban scenes from 10 video sequences captured in 4 cities in China. It includes semantic annotations for 6 detection classes and comprises of 160 training and 45 validation images. The UAV123 \cite{mueller2016benchmark} dataset includes 123 aerial video sequences, with bounding box annotations provided for every sequence. The VizDrone \cite{vizdrone} dataset has 288 video clips and 10,209 static images captured from drone-mounted cameras, manually annotated with over 2.6 million bounding boxes. The OpenDD \cite{breuer2020opendd} dataset covers seven roundabouts in Germany, including over 80,000 road bounding boxes in 62+ hours of data. The UAVDT \cite{du2018unmanned} dataset includes 100 video sequences with approximately 80,000 representative frames, captured at various urban locations such as highways and crossings.

\subsection{Event-based Camera Datasets}
Early event-based datasets have been created by converting frame-based datasets to event format for benchmarking tasks \cite{nmnist, mnist-dvs, Caltech101, cifar10-dvs}. The benefit of these methods is that large datasets can be generated without manual labeling. However, the low screen refresh rate of displays create unnatural event sequences. As a result, recent efforts have centered on creating real-world datasets for detection and recognition of persons, animals, objects, robots, and quadrotors \cite{Amir2017, postures-dvs, sl-animals-dvs, poker-dvs, pred18, iaboni2021event, iaboni2022event}. The majority of these event-based datasets have been collected in controlled, indoor environments.

\subsubsection{Real-World Urban Event Datasets}

The following discussions highlight the fact that only a handful of studies have provided open-source image/video datasets capturing urban settings \cite{scenes-dvs, miao2019neuromorphic, sironi2018hats, de2020large, perot2020learning, li2023sodformer, el2022high, li2019event, binas2017ddd17}. 

\begin{itemize}
    \item The Scenes-DVS \cite{scenes-dvs} set consists of recordings of long hikes in urban environments, with sequences of indoor and outdoor environments exceeding 15 minutes long split into small 50ms sequences. Scene classification accuracy was evaluated with a frame based single-layer and three multi-layer SNN configurations (with varying hidden layer sizes) \cite{negri2018scene}. The highest accuracy reported for the single-layer architecture and multi-layer architecture was 81.2\% and 84.4\%, respectively.
    
    \item The PAFBenchmark pedestrian detection dataset \cite{miao2019neuromorphic} contains 4670 frame images at 20ms time intervals with bounding box annotations of scenarios seen in traffic surveillance tasks such as pedestrian overlapping, occlusion, and collision. 

    \item EV-IMO \cite{mitrokhin2019ev} contains 32 minutes of multiple objects moving independently against various backgrounds in an indoor setting. Imagery was obtained at a capture speed of 200 frames per second. Annotations were provided in the form of pixel-wise masks, depth maps, and camera trajectories.

    \item The DDD17 dataset \cite{binas2017ddd17} focuses on driving scenarios to provide over 400GB and 12 hours of frame and event data captured using a 346×260 pixel DAVIS sensor. Manual annotations are provided for synchronized frame and event streams. The dataset was tested on an object detection task using a fine-tuned YOLOv3 model.
    
    \item N-Cars \cite{sironi2018hats} is a large, real-world event-based dataset for car classification. It is composed of 12,336 car samples and 11,693 non-car samples (background). Each sample lasts 100ms. Classification accuracy for the two-class problem was evaluated using a linear SVM classifier. 90\% classification accuracy was reported for this evaluation.

    \item The Gen1 Automotive \cite{de2020large} contains 39 hours of open road and various driving scenarios ranging from urban, highway, suburbs, and countryside scenes. Manual bounding box annotations are provided for two classes of detection: pedestrain and cars. The capture resolution of this set was 304 $\times$ 240. The data has not been evaluated for classification or detection accuracy in this study.

    \item The 1 Megapixel Automotive \cite{perot2020learning} dataset includes 14 hours of annotated vehicles and pedestrians captured within roadway scenes. This set is the highest resolution neuromorphic object detection dataset from a vehicle-mounted perspective to date, with a capture resolution of 1280 $\times$ 1024. A recurrent ConvLSTM architecture was used to evaluate the proposed dataset \cite{shi2015convolutional}. A mAP of 0.43 was reported for automotive detection using the proposed architecture. 

    \item The PKU-DAVIS-SOD dataset \cite{li2023sodformer} dataset contains 220 event and frame-based driving sequences and labels at a frequency of 25 Hz and a resolution of 346 $\times$ 260. The set has 276,000 labeled frames and 1,080,100 bounding boxes. A spatiotemporal Transformer architecture was used to evaluate the proposed dataset for three classes of detection.

    \item The authors of \cite{el2022high} proposed a dataset combining event- and frame-based traffic data. Using a DAVIS 240c camera operating at a resolution of 240 $\times$ 180 pixels and 24 Hz, two distinct scenes are captured with a total of 9891 vehicle annotations. A static camera was mounted to the side of a building to observe a street scene. The fidelity of the dataset was evaluated on a multi-object tracking task.

\end{itemize}

\subsection{NU-AIR Versus Other Datasets}

The proposed NU-AIR dataset represents a significant advancement over previously reported datasets by offering several distinct characteristics. Contrasting with existing benchmarks, NU-AIR capitalizes on the high temporal resolution of neuromorphic sensors, gathering data via cameras mounted on quadrotors navigating intricate urban topologies. This configuration inherently yields high-dimensional scene variations, encapsulating temporal dynamics and urban microstructures, resulting in a dataset that is both granular and temporally rich \cite{gupta2022monitoring}. Unconstrained by terrestrial limitations, drones maneuver seamlessly over intricate and demanding terrains, sidestepping challenges endemic to vehicle-based systems \cite{daknama2017vehicle, bock2020ind}. Their ability to hover at varied altitudes and unrestricted yaw, pitch, and roll orientations allows for a dataset which meticulously captures the geometric and radiometric intricacies of the surveyed environments from diverse optical perspectives \cite{marathe2019leveraging, joyce2018principles}. Conversely, vehicle-affixed imaging modalities, bound to their terrestrial constraints, primarily capture data through a restricted lateral optical aperture \cite{koch2016outdoor}. Such a perspective is inherently prone to occlusions, generated by terrain variations and physical obstacles \cite{anderson2013lightweight, puri2005survey, del2021unmanned}. The outcome is data that may be riddled with occlusion gaps and potential distortions, particularly when documenting elevated structures or expansive regions \cite{barnas2019comparison}. This stark distinction underscores the significant advantages that drones present in the landscape of spatial data acquisition \cite{liang2019drone, roldan2021survey}.

Anchoring the dataset is the integration of precise bounding box annotations, optimized for a multi-class detection framework. These annotations provide a structured ground truth that enables algorithmic training, thus facilitating high-precision performance metrics and evaluation benchmarks. By embracing the event-driven, sparse, asynchronous nature intrinsic to neuromorphic cameras, the dataset brings forth a myriad of challenges—ranging from handling high temporal discrepancies, mitigating drone-induced perturbations such as motion blur, to addressing complex visual phenomena like parallax and rapid perspectival alterations. 

NU-AIR’s heterogeneity is enhanced by diverse illumination conditions and varied environmental scenarios, including university campuses, pedestrian-centric walkways, and  dynamic traffic intersections. This diversity serves as a robust platform for advancing event-driven vision algorithms, necessitating the exploration of adaptive thresholding, noise filtering, and temporal clustering techniques to each unique stimulus.


\begin{figure}[b!]
    \centering
    \includegraphics[scale = 0.6]{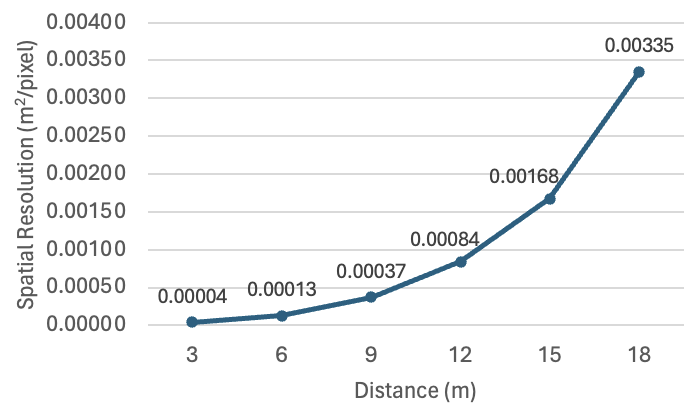}
    \caption{Spatial resolution of the camera as a function of distance, measured over a range of 3.05 to 18.29 meters.}
    \label{fig:spatial_res}
\end{figure}

\begin{figure}[t!]
    \centering
    \includegraphics[scale = 0.23]{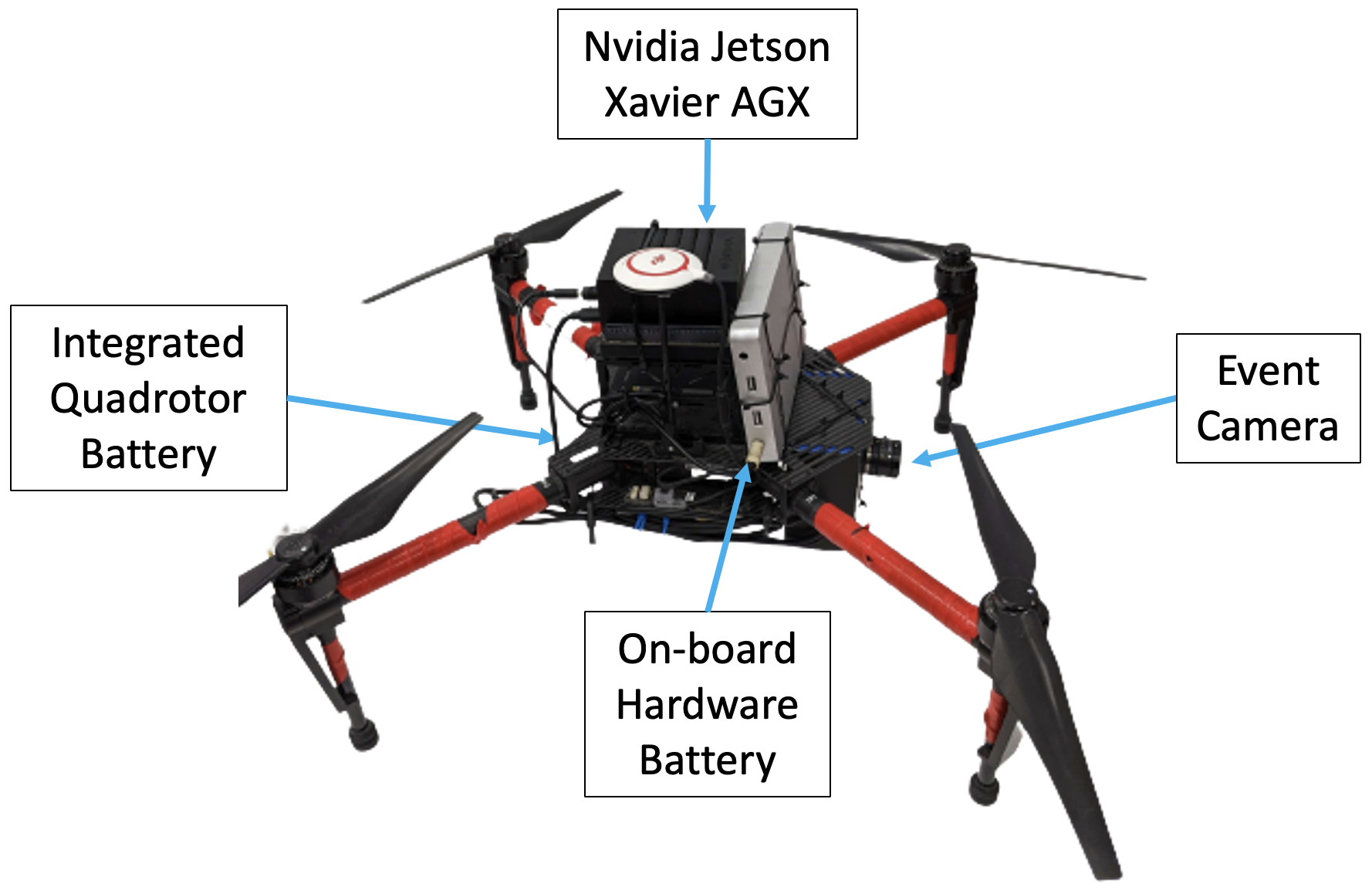}
    \caption{The DJI Matrice M100 quadrotor with a forward-facing Prophesee Gen3.1 VGA event camera was used for data collection.}
    \label{fig:quad_setup}
\end{figure}


\begin{figure*}[t!]
    \centering
    \includegraphics[angle=90, scale = .7]{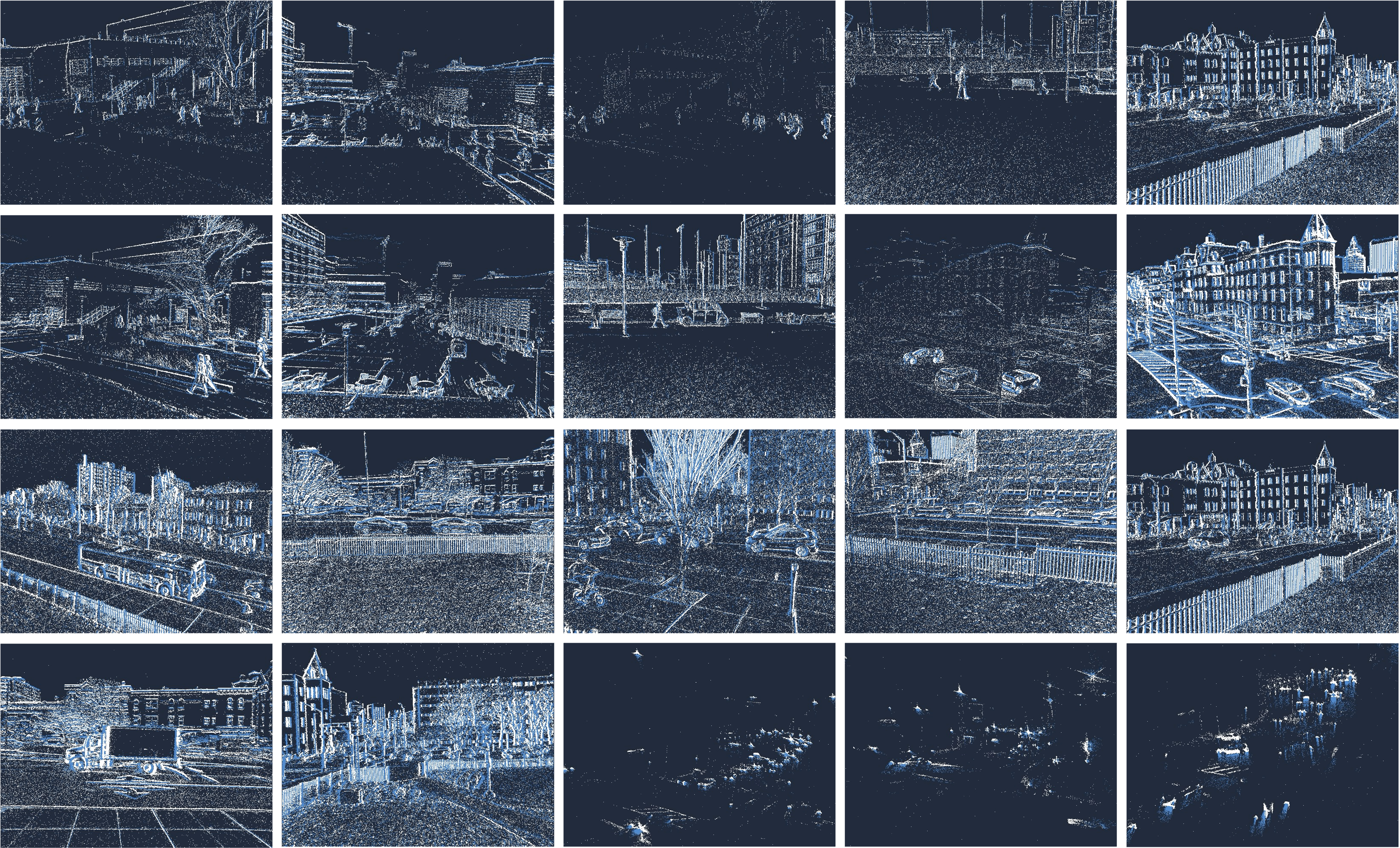}
    \caption{Sample images from the NU-AIR dataset showcasing the diversity of urban environments and the intricate challenges presented by drone-based neuromorphic imaging.}
    \label{fig_5x5}
\end{figure*}

\begin{figure*}[t!]
    \centering
    \includegraphics[scale = 0.35]{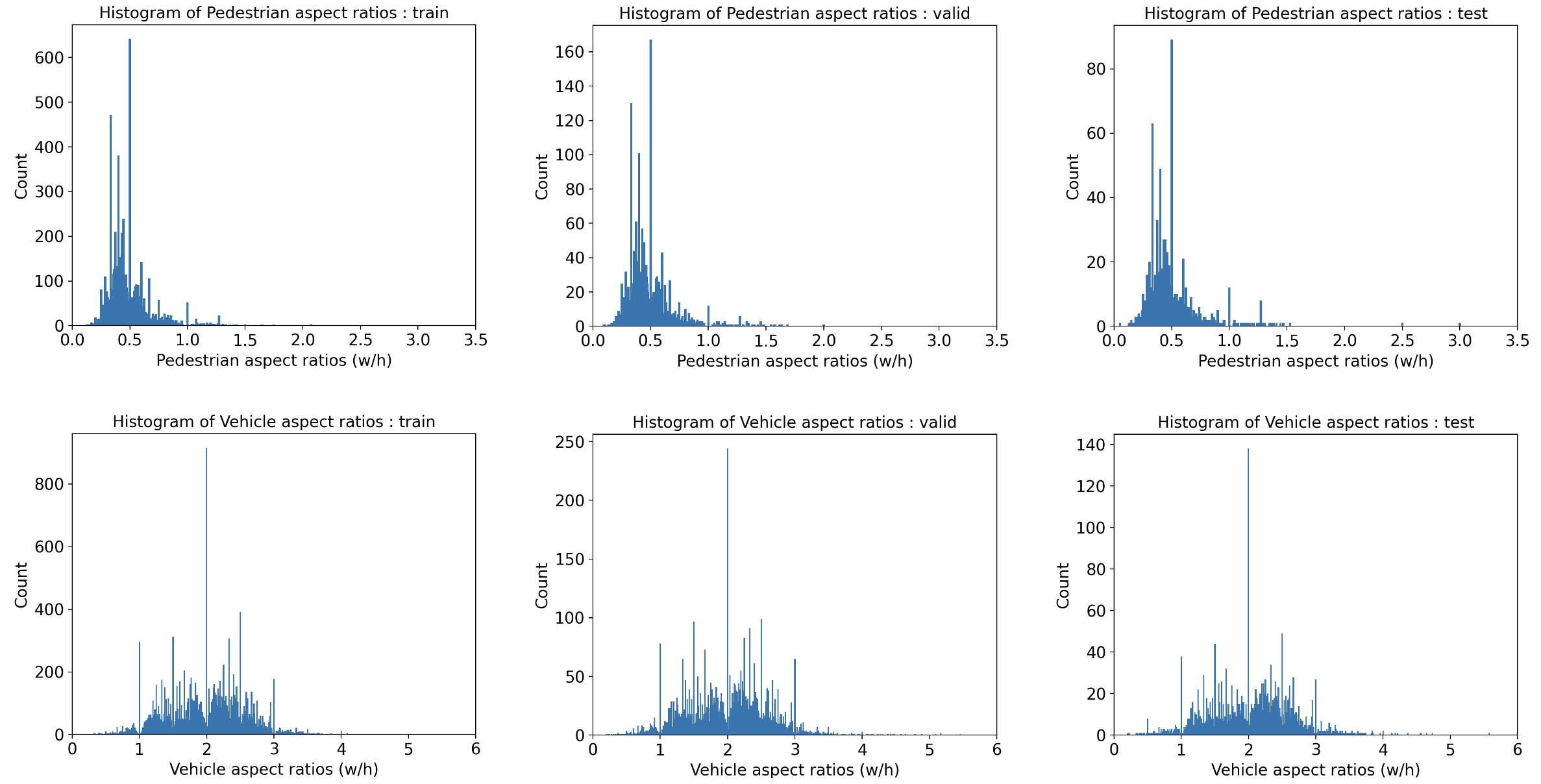}
    \caption{The ratio of width to height, known as the aspect ratio, is displayed for the manually labeled bounding boxes of pedestrians (top) and vehicles (bottom) in the dataset. The histograms for training, validation, and testing subsets of pedestrian and vehicle subjects are depicted.
    }
    \label{fig:ar_histo}
\end{figure*}

\begin{figure*}[t!]
    \centering
    \includegraphics[scale = 0.35]{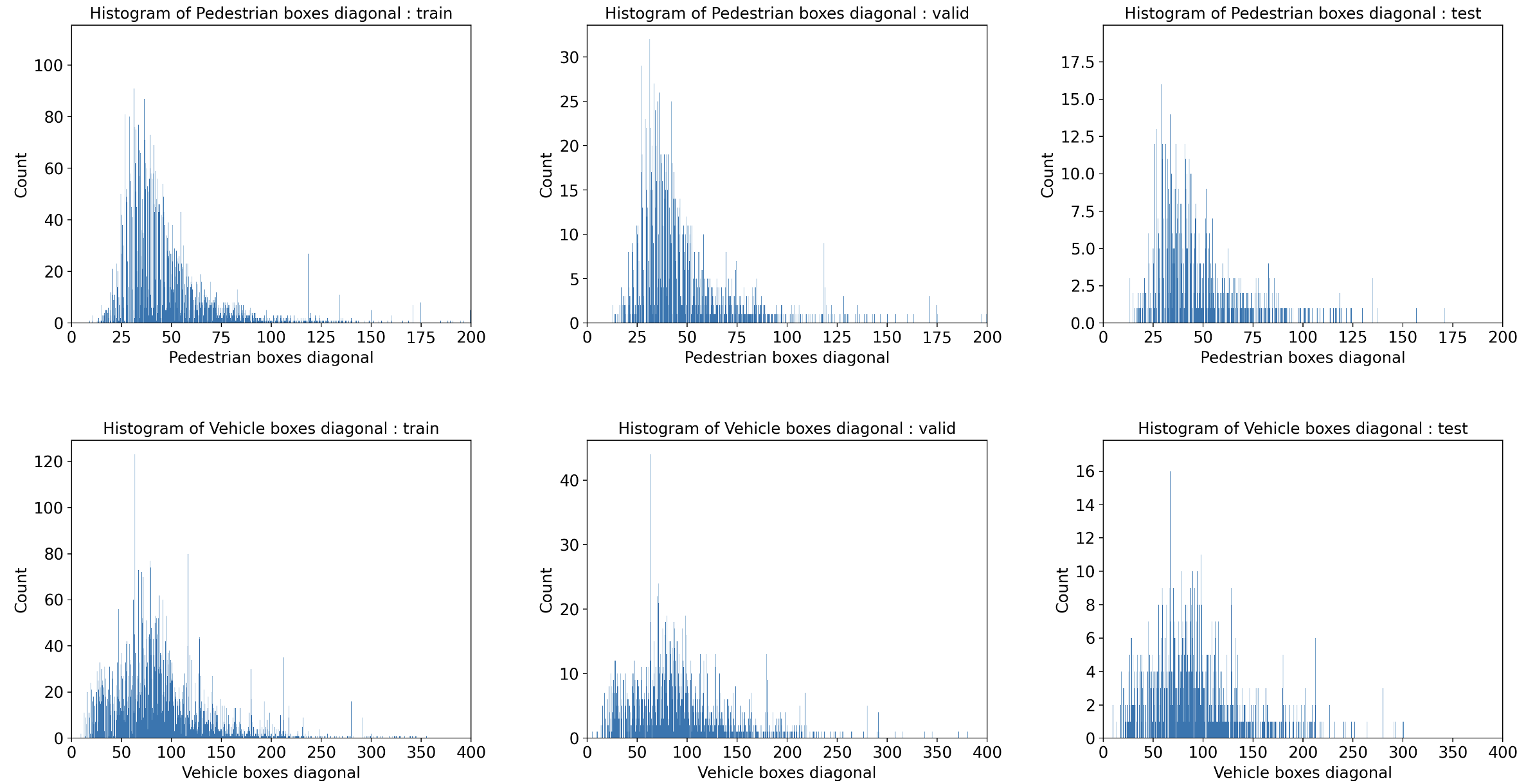}
    \caption{The diagonal distance is displayed for the manually labeled bounding boxes of pedestrians (top) and vehicles (bottom) in the dataset. Objects smaller than 10 pixels have been ignored in the annotation process. The statistics for training, validation, and testing subsets of pedestrian and vehicle subjects are depicted. 
    }
    \label{fig:dist_histo}
\end{figure*}

\section{Dataset}
\label{sec:dataset}
This section explains how the NU-AIR dataset was collected (subsection \ref{sec:record_setup}), the annotation process (subsection \ref{sec:labeling}), the file structure (subsection \ref{sec:files}), and the Voxel Cube encoding format (subsection \ref{sec:voxel}).

\subsection{Recording Setup}
\label{sec:record_setup}

This study uses a Prophesee Gen3.1 VGA (640 $\times$ 480) resolution event camera \cite{prophesee-params}. The spatial resolution of the camera measured in a controlled environment between $\thickapprox$ 3 meters and $\thickapprox$ 18 meters is depicted in Figure \ref{fig:spatial_res}. This camera operates on 1.8 V supplied via USB, with a 10 $mW$ power dissipation rating in low-power mode. The camera is equipped with a C-mount lens that offers a 70-degree field of view. The brightness intensity measurements from the sensor are used to create images at a frequency of 30 Hz, which are then manually annotated by human subjects to generate ground-truth bounding boxes around areas of interest.

As shown in Figure \ref{fig:quad_setup}, the camera is mounted on a DJI M100 quadrotor with a $10^{\circ}$ downward mounting angle \cite{DJIM100}. The camera is positioned within a custom enclosure and cushioned with anti-vibration foam with 1-inch thickness to reduce the impact of vibrations while in-flight. The camera is connected to an on-board NVidia Jetson Xavier AGX computer for data recording via USB \cite{xavier}. The recording setup and quadrotor are powered by isolated batteries to ensure reliable operations of the entire setup. The Jetson is powered by a 28800 mAh capacity DC 12V power bank, while the quadrotor is equipped with a 5700 mAh capacity TB48D battery \cite{matricebattery}. A safety rope, as shown in Figure \ref{fig:event-images}, is attached to the quadrotor for all flight tests.

The data was collected from three different real-world scenes - a crowded university campus center, a walking path with low foot traffic, and a city intersection with high vehicle traffic in New Jersey, USA. The quadrotor was translated (roll and pitch) and rotated (using yaw control) while collecting the footage. During the flights, the quadrotor operated within a speed range of 0 to 15 m/s, and at altitudes between 10 to 50 meters. The recordings range from one to five minutes in duration and were captured over the course of one week in December, at varying times of the day to include different lighting and weather conditions. A total of 70.75 minutes, split among 14 recordings, were collected, yielding 238GB of raw event data. In comparison, a frame-based camera with the same resolution and operating at a frequency of 120fps (i.e. 100 times lower temporal resolution compared to the event camera) would have generated over 359GB of data (ignoring compression and assuming 1 byte per pixel). 
Figure \ref{fig_5x5} depicts several snapshots of the acquired footage. 

Neuromorphic cameras are designed based on principles markedly distinct from those of traditional cameras. In conventional cameras, the aperture - the opening through which light passes before reaching the image sensor - is a critical component, and its size is adjustable to control the amount of light that enters. This mechanism is key for managing exposure and depth of field in photography and videography. On the other hand, neuromorphic cameras differ operationally, where each pixel in the array operates independently and asynchronously, detecting changes in light intensity at the pixel level. Rather than a physical aperture, the ‘effective’ aperture of a neuromorphic sensor is ingrained within the design of the pixel circuits themselves. This includes parameters like pixel size and contrast sensitivity thresholds, all of which determine the conditions under which a pixel will trigger an event based on changes in light intensity. Essentially, the electronic configuration of the individual receptive pixels controls the sensitivity, not a mechanical aperture. The published data sheet of camera parameters can be found at the following manufacturer URL \cite{prophesee-params}.

\subsection{Manual Annotation}
\label{sec:labeling}
By using the change detection events provided by the neuromorphic sensor, it is possible to generate images at any desired temporal resolution, precisely aligned with the events stream. Such alignment ensures that annotations on the images can be applied directly as ground truth for the events stream without requiring calibration or correction, as they share the same pixel array \cite{sironi2018hats}. 

Given the high temporal resolution of the event camera, images are generated at a frequency of 30Hz for annotation. All events within this window are considered simultaneously occurring. These aggregated events are then plotted onto a frame to create an image. In this visualization, white pixels denote areas of increased light intensity (positive events), while blue pixels highlight areas of decreased light intensity (negative events). Black pixels signify locations where no change in light intensity was observed. The images are then given to human annotators to draw bounding boxes around pedestrians and vehicles. The traditional method of drawing a bounding box, used in the annotation of datasets such as ILSVRC \cite{russakovsky2015} and introduced by Su et al. \cite{su2012}, involves manually selecting imaginary corners to tightly encase an object. This can be challenging, as the desired corners may fall outside the physical bounds of the object, often necessitating numerous adjustments to achieve an accurate bounding box. For this dataset, we opt to use the extreme clicking technique \cite{papadopoulos2017} for all bounding boxes. This method simplifies the task by requiring annotators to identify four physical points on the object: the top, bottom, left-, and right-most points. These tangible points facilitate a more natural task that is more readily performed, thus enhancing both efficiency and precision.

To ensure robust annotation fidelity, we employ a temporally coherent annotation pipeline. Specifically, annotations are propagated sequentially across frames, with the current frame inheriting annotations from the preceding timestamp. These inherited annotations are then updated and refined based on the objects present in the current frame. Annotators are instructed to leverage not just the current frame, but also the immediately preceding and succeeding frames in the sequence. This temporal context allows for a more complete interpretation of object dynamics that may be partially sparse or ambiguous in any given individual frame. Our sequential annotation approach confers two key advantages: (1) it promotes high temporal consistency in the annotation trajectory, and (2) it substantially reduces annotation overhead by leveraging the temporal redundancy and continuity across consecutive frames.


Specific instructions are provided to the annotators to minimize ambiguity and inconsistencies among annotations. Objects smaller than 10 pixels or those partially occluded (less than 50\% visible) are disregarded. Various large vehicles, such as cars, buses, and trucks, are annotated uniformly within the vehicle category. Persons within cars or buildings are excluded from annotation. A total of 66,958 vehicle and 26,246 pedestrians bounding boxes are annotated using the extreme clicking technique. For vehicles, the training set has 46,870 bounding boxes, followed by 13,391 boxes for the validation set and 6697 boxes for the test set. For pedestrians, the number of bounding boxes is 18,372 (training), 5249 (validation), and 2625 (testing). For the city intersection footage, there are 60,833 vehicular bounding boxes and 4852 pedestrian bounding boxes. For the campus center footage, there are no vehicle and 16,271 pedestrian bounding boxes. Finally, for the walking path footage, there are 6125 vehicular bounding boxes and 5123 pedestrian bounding boxes. The annotations produced were reviewed by two experts to ensure location accuracy and subject correctness. Example images with manual annotations are shown in Figure \ref{fig:event-images}.

\subsection{Dataset Format}
\label{sec:files}
The dataset was captured across 14 continuous recording sessions. In order to facilitate the training, the continuous recordings are divided into 15 seconds intervals, resulting in a total of 283 samples. The total samples are divided into training, validation, and testing subsets with 197, 57, and 29 samples in each subset, respectively. To avoid overlap between subsets, events from each continuous recording session are assigned to the same subset. The samples have been further categorized based on the time of day and the nature of the recording environments. Specifically, out of the 283 samples, 236 samples were captured during the day, while 47 were recorded at night.

Each sample was provided in a binary .dat format, where events re encoded using 4 bytes for the timestamps and 4 bytes for the position and the polarity. Specifically, $14$ bits are used for the $x$ position, $14$ bits for the $y$ position, and $1$ bit for the polarity. Python code to generate images from binary files is provided with the dataset.

Bounding box annotations follow a {\tt{numpy}} format \cite{numpy}. Each {\tt{numpy}} array contains the following fields:
\begin{itemize}
    \item $t_{s}$, timestamp of the box in microseconds
    \item $x$, abscissa of the top left corner in pixels
    \item $y$, ordinate of the top left corner in pixels
    \item $w$, width of the box in pixels
    \item $h$, height of the box in pixels
    \item ${class}_{{id}}$, class of the object: 0 for pedestrians and 1 for vehicles
\end{itemize}

Python code to load and view samples from the dataset with the corresponding annotations visualized has been provided. Within the released code repository, an example on how to apply the evaluation metrics on the dataset is provided.

\subsection{Dataset Descriptive Statistics}
\label{sec:data_eval}

The histograms displaying the aspect ratios of bounding boxes for pedestrians and vehicles are presented in Figure \ref{fig:ar_histo}. These histograms have been separately calculated for the training, validation, and testing subsets. The Python code to generate the histograms has been open-sourced for readers. It was noted that the statistics for the training, validation, and testing subsets were consistent across the different subsets.

For bounding boxes enclosing pedestrians, the mean aspect ratio is found to be 0.45. Conversely, for those enclosing vehicles, a mean aspect ratio of 2.02 is noted. These aspect ratios reflect the consistent perspective from which the quadrotor captures these objects: pedestrians are usually seen from an oblique overhead angle, resulting in more vertical bounding boxes, whereas vehicles are typically captured from side or top views, leading to more horizontal bounding boxes.

In addition, the distribution of the diagonal distances for the bounding boxes of pedestrians and vehicles is depicted in Figure \ref{fig:dist_histo}. The mean diagonal distance for bounding boxes around pedestrians is calculated to be 47.75 pixels, and for vehicles, it is calculated to be 90.17 pixels.

\section{Performance of Baseline DNNs and SNNs}
\label{sec:eval}
In this section, experiments are provided to quantify the performance of baseline models for object detection on the NU-AIR dataset.

\subsection{Frame-based DNNs}
A total of 10 single-stage or two-stage frame-based object detection DNNs are used in this study and their COCO mean average precision (mAP) metric is reported. The models, trained from random initialization without freezing any layers, include YOLOv5n \cite{glenn_jocher_2020_4154370}, YOLOv5s \cite{glenn_jocher_2020_4154370}, YOLOv6n \cite{li2022yolov6}, YOLOv6s \cite{li2022yolov6}, YOLOv7t \cite{wang2022yolov7}, YOLOv7 \cite{wang2022yolov7}, YOLOv8n \cite{Jocher_YOLO_by_Ultralytics_2023}, YOLOv8s \cite{Jocher_YOLO_by_Ultralytics_2023}, Faster-RCNN \cite{ren2015faster}, and RetinaNet \cite{lin2017focal}. 

Training is performed on Google Colaboratory with the standard GPU and memory configuration provided by the platform \cite{colab}. All YOLO models are trained using the SGD optimizer, with a weight decay of 5$e^{-4}$ and momentum of 0.937. Each model uses a learning rate of 0.01. The training process spans 100 epochs with a batch size of 16. The Faster-RCNN model uses the SGD optimizer with a learning rate of 0.001 and momentum of 0.9, with a weight decay of 5$e^{-4}$.The RetinaNet model uses the SGD optimizer with a learning rate of 0.01 and momentum of 0.9, with a weight decay of 1$e^{-4}$.

Training of these networks is performed using images generated according to the method described in Section \ref{sec:record_setup}, where all events within a 33.33 ms time window are treated as if they occur simultaneously. These collective events are then visualized on a single image, with white pixels representing areas with an increase in light intensity (positive events), blue pixels representing areas with a decrease in light intensity (negative events), and black pixels representing locations with no change in light intensity.

The evaluation involves selecting the two smallest configurations of each variant of the YOLO object detector network, except for YOLOv7, where the smallest two networks are named YOLOv7-tiny and YOLOv7. The network size selection is made to evaluate the networks' performances on edge devices that compete with low-power neuromorphic hardware. For Faster-RCNN and RetinaNet, only one variant each is evaluated. The number of parameters in each model is reported in millions. 

\subsubsection{DNN Results}
Table \ref{tab_anns} depicts the performance of various DNNs for the NU-AIR dataset. From the table, it is evident that YOLOv7 achieved the highest mAP of 0.53, whereas both RetinaNet and YOLOv5n reported an identical mAP of 0.37, which was among the lowest scores. Additionally, the runtime is reported alongside these performance metrics to provide a comprehenisive view of how each model operates under the NU-AIR dataset conditions. Runtime, in this context, encompasses the total time taken for the model to process input data and output a result, which includes not only the inference time but also pre-processing and post-processing stages. 

The variations in runtime across different DNN models provides critical insights for deploying of these models in practical settings, reflecting the inherent trade-offs between model complexity and efficiency. The YOLO variants exhibit a spectrum of runtimes, with smaller models like YOLOv5n and YOLOv6n expected to offer faster inference times due to their reduced size and simpler architectures. Conversely, larger models like YOLOv7, while delivering a higher mAP score, also exhibit longer runtime. Additionally, the runtimes of models such as Faster R-CNN and RetinaNet highlight the impact of architetural differences on processing speed. Faster R-CNN, with it's two-stage detection process, shows significantly longer runtime despite its effectiveness in object detetion. RetinaNet, offering a middle ground in terms of speed, still lags behind the faster YOLO models.

Despite the general trend observed where larger models like YOLOv7 perform better than smaller models, there are anomalies where smaller models manage to outperform or closely match their larger counterparts. For instance, the relatively smaller YOLOv5s surpasses the performance of the considerably larger RetinaNet. Additionally, the smaller YOLOv6 variant does follow the expected pattern, and outperforms its larger sibling. This observation, although indicative of the nuances within specific model architectures and their adaptability to the dataset, still aligns with the broader trend in deep learning. 


\begin{table}[t!]
\centering
\caption{The network size in millions of parameters (Par.), COCO mean average precision (mAP), and average runtime (ms) for 10 object detection models trained on the NU-AIR dataset. Runtime (RT) results were computed as the mean of three separate assessments. The results presented below were obtained by evaluating each model on the dataset's test set.}
\begin{tabular}{|l|l|l|l|}
\hline
\textbf{DNN Methods} & \textbf{Par.} & \textbf{mAP} & \textbf{RT} \\ \hline
YOLOv5n          & 1.9M            & 0.37     &       31             \\ \hline
YOLOv5s          & 7.2M            & 0.39     &       38             \\ \hline
YOLOv6n          & 4.3M            & 0.41     &       32             \\ \hline
YOLOv6s          & 17.2M           & 0.39     &       41             \\ \hline
YOLOv7t          & 6.2M            & 0.36     &       64             \\ \hline
YOLOv7           & 36.9M           & 0.53     &       58             \\ \hline
YOLOv8n          & 3.2M            & 0.42     &       28             \\ \hline
YOLOv8s          & 11.2M           & 0.44     &       47             \\ \hline
Faster-RCNN      & 19M             & 0.42     &       117             \\ \hline
RetinaNet        & 36.4M           & 0.37     &       83              \\ \hline
\end{tabular}

\label{tab_anns}
\end{table}

\subsection{Event-based SNNs}

\label{sec:impl_det}
Spiking Neural Networks (SNNs) are inspired by the discrete and asynchronous nature of biological neuron communication. Their compatibility with event-based data makes them particularly suited for processing information from event cameras. SNNs are being increasingly used for tasks such as object detection, motion tracking, and sensor fusion in conjunction with event cameras \cite{ Escobar2009ActionRU, kim2019spikingyolo}. Three SNNs are trained from random initialization without freezing any layers and their performance is evaluated in terms of accuracy. All SNN models are trained using the Decoupled Weight Decay Regularization (AdamW) optimizer \cite{loshchilov2019decoupled}, with a weight decay of 1$e^{-4}$. Each model starts with an initial learning rate of 1$e^{-3}$. The training process, spanning 50 epochs with a batch size of 64, is executed on a Google Cloud Compute instance equipped with NVidia Tesla T4 GPU and 64 GB of memory \cite{gcp}. A cosine annealing scheduler is used to gradually reduce the learning rate to 0. Convolutions are initialized with the Kaiming uniform method \footnote{He, Kaiming, et al. "Delving deep into rectifiers: Surpassing human-level performance on imagenet classification." Proc. IEEE ICCV. 2015}. Batch normalization layers began with a weight of 1 and a bias of 0. 

\subsubsection{Voxel Cube Encoding}
\label{sec:voxel}
Voxel cubes were used to encode event data for SNN training \cite{Cordone_2022_IJCNN}. This approach offers advantages over event encoding strategies such as event frames, two-channel images, and time surfaces in that events are not collapsed to a 2D grid, thus conserving both binarity and temporal information while allowing for batched processing of events \cite{gallego2020event}. The voxel cube encoding method is inspired by voxel grid encoding. Voxel grid encoding is an encoding method in which a series of bins are used to convert the temporal dimension to a set of event volumes \cite{bardow2016simultaneous}. A voxel grid is represented by a 3D THW tensor of events, where T is the number of timesteps, and H and W are the height and width of the data respectively \cite{bardow2016simultaneous}. The accumulation of events within the time interval is either a sum \cite{fang2021incorporating} or a binary accumulation \cite{cordone2021learning}. 

In the case of binary accumulation, when events within the represented time interval do not overlap on a particular voxel, the complete set of events can be retrieved from the representation. When multiple events intersect on a voxel, only the most recent event is represented in the voxel. Thus, temporal information describing the frequency of the event occurrence at the given voxel is lost, but the volume still reflects the distribution of events across spatial and temporal dimensions. 

By contrast, the voxel cube encoding technique subdivides each event volume into smaller sub-volumes, referred to as voxel cubes. A voxel cube is represented by a 4D CTHW tensor, where C is the number of channels, T is the number of timesteps, and H and W are the height and width of the data respectively \cite{Cordone_2022_IJCNN}. The events within a sub-volume are assigned to a cell depending on position (x,y) and to a sub-volume depending on timestamp (t). Rather than counting event frequency like a histogram, each event is counted based on its temporal distance from the center of neighboring sub-volumes.
The value of each event is distributed over the two nearest sub-volumes, and weighted by the temporal distance between the event and the centroids of neighboring sub-volumes. In this way, temporal information is preserved as all events that fall within a sub-volume give a varying contribution based on the event timestamp. In this work, T, the number of timestamps, corresponds to 5.

\subsubsection{Network Details}
The Parametric Leaky Integrate-and-Fire (PLIF) neurons integrated specific parameters to expand the adaptability and versatility of the network. The PLIF neuron is a spiking neuron with learnable membrane parameters, thus enhancing the expressiveness of the SNN \cite{fang2021incorporating}. The time constant of the neuron's membrane potential, represented by the parameter $\tau$, is set to a default value of 2, determining the rate at which the potential decays towards its resting state. When the accumulated potential of the neuron reaches or surpasses the set threshold of 1, the neuron emits a spike. To allow gradient computation during the backward pass, especially given the non-differentiability of the actual spiking function, the ATan (arctangent) function is used as a surrogate, offering a smooth and differentiable approximation. Gradient norms feature an upper-bound of 1 to prevent exploding gradients. The final results are obtained by averaging over three training runs. The SNNs featured the following details: 

\begin{enumerate}
    \item {\it Spiking VGG \cite{simonyan2014very}:} VGG (Visual Geometry Group) is a CNN that consists of up to 19 convolutional layers, followed by 3 fully-connected layers. The spiking variant of VGG includes batch normalization layers added before each spiking convolutional layer. The VGG SNN in this work features 11 convolutional layers.
    
    \item {\it Spiking DenseNet \cite{huang2017densely}:} DenseNet is an architecture that promotes gradient propagation by using channel-wise concatenations, which is an operation that preserves spike representation. DenseNet configurations are defined by the depth (number of layers within the network) and growth rate (how much new information each layer contributes to the global state). For the spiking variant, ReLU activations are replaced with PLIF. The spiking DenseNet architecture in this work uses a depth of 120 convolution layers and 1 fully connected layer with a growth rate of 24.
    
    \item {\it Spiking MobileNet \cite{howard2017mobilenets}:} MobileNet is a model intended for use in mobile applications that use depth-wise separable convolutions which need fewer parameters and computations compared to traditional convolutions. By varying the number of input channels, MobileNet can be evaluated at various sizes. The spiking version of MobileNet removes the activation function between the depth-wise and point-wise convolutions, and positions all batch normalization layers before the convolutional layers. The spiking MobileNet in this work uses 16 input channels.
\end{enumerate}


The SNNs are combined with Single Shot MultiBox Detector (SSD) detection heads to perform object detection. The SSD object detection framework consists of a spiking classifier backbone and multiple predictor heads \cite{liu2016ssd}. Each head is responsible for two operations within the detection pipeline. The first operation involves bounding box regression, wherein each head projects potential object locations at varied scales and aspect ratios. Simultaneously, in a second operation, these heads attribute class probabilities to each delineated bounding box. Readers are referred to \cite{Cordone_2022_IJCNN} for further implementation details about the SNNs used for the analysis presented here.

For the task of object detection, the mAP over 10 IoU ([0.5:0.05:0.95] -- the COCO mAP) is reported \cite{lin2014microsoft}. However, assessing the performance of SNNs is not limited to mAP, as multiple others features are needed to take advantage of their benefits when embedded in specialized hardware. For all results, the following metrics are reported in Table \ref{tab_spikeresults}:
\begin{itemize}
    \item {\it Number of parameters:} To meet the high memory constraints of embedded systems, the network size is measured in terms of the number of parameters.
    
    \item {\it Number of accumulation operations (ACCs) :} SNNs do not require multiplicative operations, enabling substantial energy savings on specialized hardware. The number of accumulation operations (ACCs) is reported to highlight potential energy savings. All spiking convolution operations amount to ACCs, and each PLIF neuron only requires 1 ACC per timestep to update its potential. Batch normalization layers are not included in the ACCs count.
    
    \item {\it Sparsity:} The number of spikes emitted after each activation layer is measured to represent the global sparsity of the network compared to an fully dense equivalent Deep Neural Network (DNN). Processing events with SNNs preserves the data sparsity \cite{kugele2020efficient}. On specialized hardware, computations are only performed when there are spikes, therefore a highly sparse network would consume less power than its dense counterpart. The sparsity percentage for the network is quantitatively assessed by first determining the total number of spikes emitted during the forward pass, denoted as $n^{forward}_{spikes}$. This count is then normalized by the theoretical maximum number of spikes, denoted as $n_{max}$, which corresponds to the size of the output feature map. The sparsity percentage (S) for a given layer can be expressed as:
    \begin{equation}
        S = \left(1 - \frac{n^{forward}_{spikes}}{n_{max}}\right) \times 100\%
    \end{equation}
    This calculation is conducted across the entire test set.

\end{itemize}

\begin{table}[b]
\centering
\caption{Comparison of three tested SNN configurations detailing network size in millions of parameters (Par.), number of accumulations per timestep, network sparsity (Spar.), and Runtime in milliseconds (RT). Runtime results were computed as the mean of three separate assessments. The results presented below were obtained by evaluating each model on the dataset's test set.}
\begin{tabular}{|p{1.6cm}|p{.7cm}|p{1.1cm}|l|l|l|}
\hline
\textbf{SNN Methods}     & \textbf{Par.} & \textbf{ACC/ts} & \textbf{Spar.} & \textbf{mAP} & \textbf{RT} \\ \hline
VGG-11 + SSD         & 12.6M          & 11.1G          & 23.1\%  & 0.17  &  192    \\ \hline
DenseNet121-16 + SSD & 8.2M            & 2.3G           & 38.1\% & 0.17  &  247    \\ \hline
MobileNet-16 + SSD   & 24.3M          & 4.3G           & 27.9\%  & 0.13  &  166    \\ \hline
\end{tabular}
\label{tab_spikeresults}
\end{table}

\subsubsection{SNN Results}
For SNN object detection models, three configurations of classifier and SSD network are used for evaluation. The COCO mAP accuracy and runtime results of the three SNNs trained on this dataset are given in Table \ref{tab_spikeresults}.

\begin{itemize}
    \item The VGG-11 + SSD network configuration achieves a COCO mAP of 0.17, which is tied with DenseNet121-16 + SSD for the highest evaluation metric observed in this work. The size of this network configuration is 12.6M parameters, with 11.1B accumulation operations per timestep, and a sparsity of 23.1\%. The runtime performance registered at 192ms, positioned between the faster MobileNet and slower DenseNet configurations.

    \item The DenseNet121-16 + SSD network configuration achieved a COCO mAP of 0.17. The network contained 8.2M parameters. The number of accumulation operations per timestamp for this network is the least among the evaluated SNNs at 2.3B. At 38.1\% sparsity, this network is the most dense of the evaluated SNNs. This configuration also exhibits the longest runtime at 247ms.

    \item The MobileNet-16 + SSD network configuration achieved a COCO mAP of 0.13. This is the lowest reported mAP result among the three SNNs used for dataset evaluation. This network configuration is the largest of the evaluation SNNs at 24.3M parameters. This network performs 4.3B accumulation operations per timestep, with a sparsity of 27.9\%. Notably, the runtime performance is recorded at 166ms, making it the fastest among the models reviewed.
\end{itemize}

\begin{figure*}[t!]
    \centering
    \includegraphics[scale = .65]{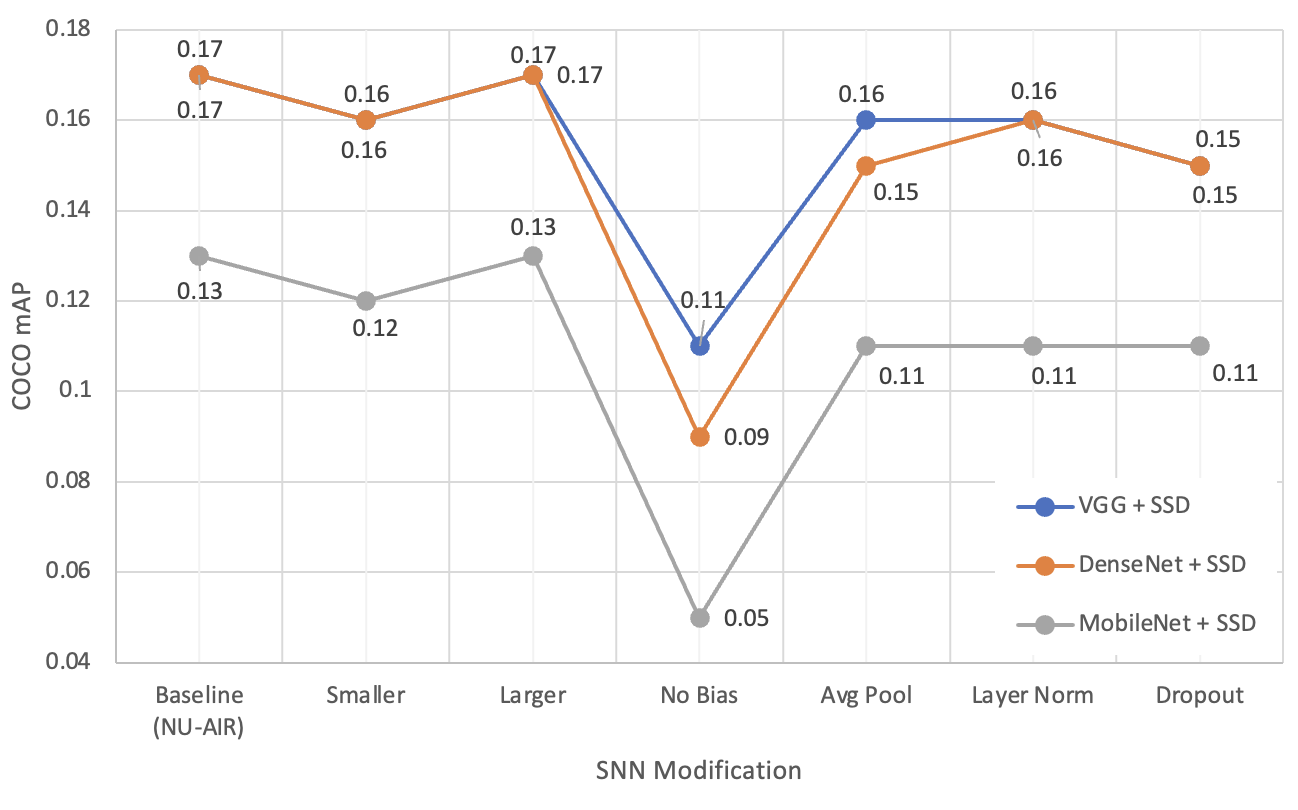}
    \caption{Comparison of the COCO mean average precision (mAP) for three spiking neural networks (SNNs) and their ablations trained on the presented dataset. 
    }
    \label{fig:ablat}
\end{figure*}

For comparison, the SNN results achieved above are benchmarked against the COCO mAP results achieved in \cite{Cordone_2022_IJCNN} and are depicted in Figure \ref{fig:ablat}. The authors in \cite{Cordone_2022_IJCNN} trained their model on the Prophesee GEN1 dataset. It is observed that the SNNs trained on the NU-AIR dataset achieve mAP close to the mAP of the comparison study \cite{Cordone_2022_IJCNN}. The COCO mAP of the VGG-11 + SSD model reached 0.17, and is the highest accuracy reported of the three trained models. However, \cite{Cordone_2022_IJCNN} reported that DenseNet121-24 + SSD model achieved the highest accuracy in their work. 

In terms of mAP, the VGG-11 + SSD and DenseNet121-16 configurations reports a higher object detection performance than the MobileNet-16 + SSD configuration. The DenseNet121-16 configuration achieves the mAP score with a smaller network and less accumulations than the VGG-11 architecture. This trade-off in performance may be valuable on neuromorphic systems with constrained amounts of neurons. However, VGG-11 uses the most sparse network structure, which is best suited for use in power constrained embedded hardware. 

The lower performance of the MobileNet architecture was observed to be related to the use of depth-wise separable convolutions. In \cite{Cordone_2022_IJCNN}, depth-wise separable convolutions were shown to reduce performance in smaller networks (less than 32 input channels) when compared to normal convolutions. Although depth-wise separable convolutions enable the use of a smaller number of network parameters, thus making the training of the SNN with surrogate gradients a simpler task, small networks were shown to benefit from the additional network size for a 5\% increase in accuracy. Thus, a larger variant of MobileNet (larger than 16) may be used to yield an increase in network performance.

Regarding runtime, the DenseNet and VGG configurations, despite their higher accuracy, demonstrate slower processing times as compared to MobileNet. Longer runtimes could limit applicability in time-sensitive environments. Conversely, MobileNet, with a quicker runtime, stands out as the most viable option for scenarios requiring rapid response despite lower accuracy.

\subsubsection{Comparative Analysis of SNN and DNN Results}
The baseline results presented in Sections 4.1 and 4.2 reveal notable performance disparities between SNN and DNN models. While SNNs are theoretically well-suited for processing the dynamic, event-driven nature of the NU-AIR dataset, the results indicate that DNNs exhibited superior performance in terms of overall mAP. Specifically, DNNs achieve a 0.41 average mAP, significantly outperforming the SNNs, which attain a 0.16 average mAP. This performance gap underscores a critical challenge for SNNs: despite their design for temporal sequence processing, their ability to effectively encode and process spatial features presents an area ripe for in-depth exploration into optimization.

\begin{itemize}

\item 
{\bf Discussion: Static and Dynamic Objects in the Scene:} As the event camera itself is in motion due to the motion the quadrotor, the pixels respond to objects in the scene that are not moving. This unique feature of the dataset in terms of dynamic and static objects, challenges the current SNN architectures and training methodologies. Of note is the interaction between static objects, like parked cars, and the dynamic foreground they inhabit. Typically, SNNs are trained to detect temporal patterns, identifying movement and changes over time. However, the presence of static objects alongside moving objects in the scene may introduce inference challenges due to overlapping temporal signatures between static and dynamic objects of the same class. The subtle interplay between static and dynamic features invites further examination to understand how these contrasting elements affect the overall performance and decision-making process of the network.

\item 

{\bf Discussion: Low-Frequency vs High-Frequency Camera Motions}: While dampening foam is used to mitigate the impact of high-frequency vibrations arising from the quadrotor operation, it does not completely eliminate the influence of the camera's motion. The dampening foam primarily filter out high-frequency jitters, leaving behind a smooth yet non-negligible low-frequency motion. This residual motion is the consequence of the drone's minor positional adjustments required to maintain stable hovering or follow flight trajectories. While subtle, these adjustments result in continuous alignment and realignment along the x, y, and z axes of the camera's frame of reference. This motion affects the performance of DNNs and SNNs differently given that they use different representations of the event dataset. Characterizing the effects of these camera motions on the performance of different NNs  presents an avenue for future research.

\end{itemize}

\subsubsection{Evaluation of Recurrent Vision Transformer}

The Recurrent Vision Transformers (RVT) introduced for object detection with event-based cameras represent an innovative integration of Recurrent Neural Networks (RNNs) and transformers applied specifically to the unique characteristics of event-based data. This research, proposed by Gehrig et al. \cite{gehrig2023recurrent}, details a novel backbone aimed at enhancing low-latency object detection in high-dynamic environments, a significant leap given the inherent challenges with event-based vision systems. The RVT model, combining the temporal dynamics handling of RNNs with the powerful spatial feature processing of transformers, sets a modern benchmark in processing efficiency and detection accuracy for neuromorphic vision systems when applied to the NU-AIR dataset. The RVT model, specifically the base version (RVT-B), achieves a mAP of 0.36 on the NU-AIR dataset, with an inference time of $\thickapprox$ 28ms. This showcases the model's capability to provide efficient and fast object detection in real-time applications for neuromorphic vision systems.

\subsection{Ablation Study of Baseline SNNs} 

\subsubsection{Effect of Depth Variation on SNN Performance}
The network's architecture size can significantly impact the model's performance. It's often observed that larger networks perform better, mainly due to their enhanced capacity for learning complex representations \cite{brutzkus2019larger, shazeer2017outrageously}. However, this increased performance often comes with a trade-off in terms of computational overhead and the risk of overfitting \cite{srivastava2014dropout, lawrence2000overfitting}. This section presents results about the effect of changing the network size on SNN performance using variations of MobileNet, VGG, and DenseNet architectures.

For each of the three baseline models (MobileNet-16, VGG-11, and DenseNet121-16), two additional networks are trained: a smaller (MobileNet-13, VGG-9, and DenseNet-98) and a larger (MobileNet-19, VGG-16, and DenseNet-142). The smaller networks has fewer layers and parameters than the baseline models, while the larger networks has more. Performance metrics of this ablation are presented in Figure \ref{fig:ablat}.

The smaller MobileNet-13, VGG-9, and DenseNet-98 achieves a relative decrease in mAP of 8\%, 6\%, and 6\%, respectively, compared to their baseline counterparts. This performance reduction can be attributed to the decreased model capacity, leading to less effective extraction and representation of complex features in the input data. The larger network variations show no gains over the baseline models, despite their additional capacity.

The larger networks come at the expense of substantially increased computational complexity and training time, bringing forth the question on cost-to-benefit. These results indicate overfitting risks, optimization difficulties due to deeper architectures, and lack of gains from additional capacity on a finite dataset.

\subsubsection{Effect of Bias Parameter Removal on SNN Performance}

The bias parameter, usually initialized with a small constant or random value, has been recognized as a pivotal element in the structure of neural networks to provide an additional degree of freedom and enhance the ability of the network to generalize beyond the training data \cite{dongare2012introduction, geman1992neural}. This section presents an analysis of the consequences of removing bias parameters on the learning and subsequent network performance. The evaluation was performed by retraining the VGG-11, MobileNet-16, and DenseNet121-16 networks without bias parameters on the NU-AIR dataset. The removal of bias parameters is performed by removing the bias initialization argument at each layer of these networks. The mAP performance of these models is subsequently compared with their conventional counterparts (i.e., with bias parameters intact). Figure \ref{fig:ablat} depicts the bias-free training results. 
\\
Specifically, for VGG, the baseline mAP was 0.17, which decreases to 0.11 when trained without bias. For DenseNet, the mAP reduces from a baseline of 0.17 to 0.09 upon bias removal. Similarly, MobileNet experiences a decline from 0.13 to 0.05 when the bias parameters were removed.
\\
The experimental results demonstrate a notable decrease in accuracy upon the removal of bias parameters from the VGG-11, MobileNet-16, and DenseNet121-16 networks. Such performance degradation is not unexpected. The bias parameters fundamentally serve to shift the activation function, assisting the learning process by introducing an additional degree of freedom that allows for a better fit to the training data. Bias removal, therefore, is akin to constraining the model's capacity to learn intricate data distributions. Without bias parameters, the resulting linear transformations (imposed by weights alone) lack the necessary flexibility to adequately model the non-linear decision boundaries present in high-dimensional data spaces, such as those of the NU-AIR dataset. Consequently, the effective learning capacity of the network diminishes, leading to a drop in mAP performance.

\subsubsection{Effect of Dropout Layers on SNN Performance}

Dropout is a common regularization technique used to reduce overfitting by deactivating a subset of neurons during training \cite{srivastava2014dropout}. The result is that the network creates more robust internal representations and reduces the model's reliance on specific feature pathways \cite{baldi2013understanding}. This section presents an analysis of the impact of using Dropout compared to its absence in the VGG-11, MobileNet-16, and DenseNet121 architectures. A variant of each model is trained on the dataset with Dropout layers added after each pooling layer. In this work, a dropout rate of 0.5 was used.

The mAP performance of these Dropout-inclusive models is compared with the Dropout-exclusive models trained according to Section \ref{sec:impl_det}.
Figure \ref{fig:ablat} depicts the Dropout-inclusive training results. The results demonstrate that the use of Dropout results in a reduction in the mAP for the three models. Specifically, the mAP for VGG-11 decreases from 0.17 to 0.16, MobileNet-16 from 0.13 to 0.11, and DenseNet121 from 0.19 to 0.16. Deactivating neurons during training provided no tangible benefit.

The observed reduction in mAP is consistent with the function of Dropout as a regularization technique. Dropout's method of randomly disabling neurons during training, which, while aimed at preventing overfitting by fostering a more generalized model, can also strip away critical information necessary for accurate precision. The resulting models are unable to generalize effectively to unseen data, thus leading to a decrease in test set mAP performance. The results highlight the importance of Dropout in striking a balance between model complexity and generalizability, ultimately leading to better performance on the real world data.

\subsubsection{Effect of Average Pooling Layers on SNN Performance}
Pooling layers are fundamental to CNNs and SNNs, providing translation invariance and reducing computational complexity by downsampling the feature maps \cite{scherer2010evaluation, sun2017learning}. Max Pooling and Average Pooling are two commonly used strategies for spatial downsampling in SNNs \cite{nagi2011max, gholamalinezhad2020pooling}. This section presents a comparative analysis of the two pooling strategies, focusing on their effect on mAP performance of VGG-11, MobileNet-16, and Densenet121 networks trained on the NU-AIR dataset. Each network is trained under two distinct configurations: one with Max Pooling layers (baseline configuration), and another with Average Pooling Layers. The results are depicted in Figure \ref{fig:ablat}. 

The evaluation results depict a notable difference in performance across the two configurations. For VGG-11, replacing Max Pooling layers with Average Pooling results in a decline of 6\% in the mAP. This decrement is more pronounced for the MobileNet-16 and DenseNet121 networks, which see a drop of 16\% and 12\%, respectively.

These results indicate a clear preference for Max Pooling over Average Pooling in the context of the NU-AIR dataset and these specific SNN architectures. This drop in performance can be attributed to the fact that while Max Pooling captures the most dominant feature in a particular region, Average Pooling computes the mean of the features which may lead to a loss of critical discriminative information.

\subsubsection{Effect of Batch Normalization vs Layer Normalization on SNN Performance}

Normalization techniques stabilize and accelerate the learning process by regulating network inputs along either the batch \cite{ioffe2015batch, bjorck2018understanding, santurkar2018does} or feature dimensions \cite{ba2016layer}. This section discusses the effect of replacing BatchNorm with LayerNorm in VGG-11, MobileNet-16, and DenseNet121 architectures trained on the NU-AIR dataset. These networks were trained with two distinct configurations: one with BatchNorm (baseline configuration), and the other with LayerNorm. The comparative analysis was carried out using the mAP metric. 
The results are shown in Figure \ref{fig:ablat}.

The comparative analysis reveals a performance difference across the two configurations. For the DenseNet121 and VGG-11 architectures, replacing BatchNorm with LayerNorm leads to decreases of 6\% in mAP. The difference was more pronounced in the MobileNet-16 network, which experienced a drop in mAP of 16\%.

These results can be attributed to the core differences between BatchNorm and LayerNorm. BatchNorm normalizes across the batch dimension, thus leveraging batch statistics to maintain a stable distribution of activations. BatchNorm also provides implicit regularization by introducing noise to the training process. This noise is created by fluctuations of the mean and variance of the preceding layer activations across different batches. On the contrary, LayerNorm normalizes over the feature dimension, thus maintaining a consistent activation distribution within each individual sample. LayerNorm computes normalized statistics from individual samples, reducing the inherent noise as well as the model's ability to generalize beyond the training data, as reflected in the lower mAP results.

\begin{figure}[t!]
    \centering
    \includegraphics[scale = 0.36]{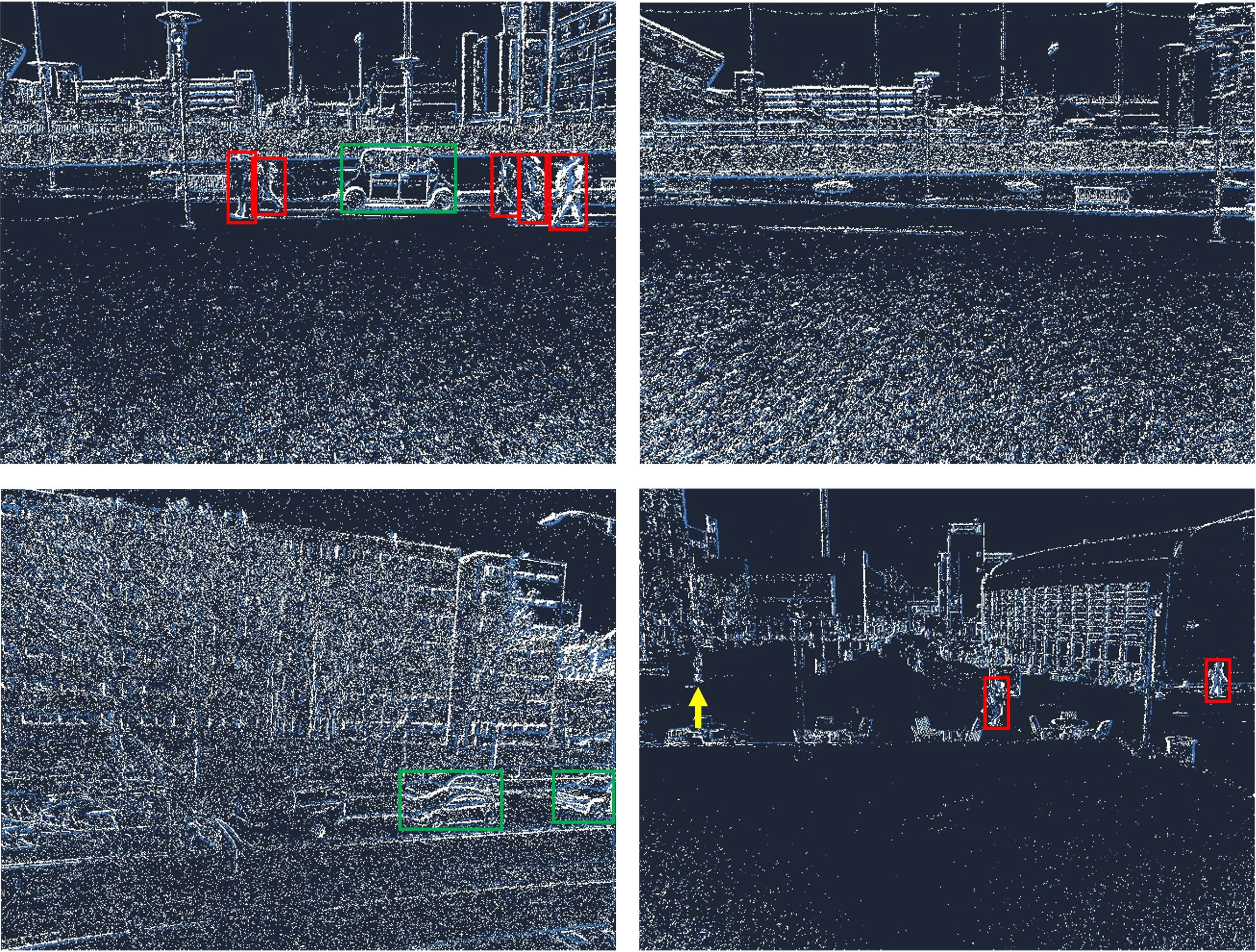}
    \caption{Success and failure cases of the baseline VGG-11 + SSD detection method on NU-AIR depicted, clockwise from top left: True Positive: Successful detection of vehicles and pedestrians; True Negative: Correct absence of detections; False Negative: Missed detection of a pedestrian (indicated by yellow arrow); False Positive: Two vehicles inaccurately identified as a single vehicle.}
    \label{fig:success_fail}
\end{figure}

\section{Limitations}
\label{sec:discuss}

Some limitations of the presented work are discussed here. 

\begin{enumerate}
    \item The SNNs evaluated in this study are trained and evaluated on GPU hardware, which differs from the low-power neuromorphic hardware that is the intended target for this type of application. Therefore, the reported performance may not be representative of how these models would perform in real-world edge devices. The recently released Intel Loihi 2 could be used to further enhance the dataset benchmarking experiments and results  \cite{orchard2021efficient}.


    \item The data used to train and evaluate the models is collected from a single city in New Jersey, which may limit the generalizability of the reported results to other settings. Future work should evaluate the performance of these models using data collected from multiple urban settings to assess the generalizability of the models.
    
    \item Aerial data acquisition through drones inherently grapples with specific drone-induced artifacts that perturb the optical fidelity of the captured sequences \cite{seifert2019influence}. These artifacts, stemming from intricate drone kinematics and atmospheric interferences, imprint stochastic translational and rotational perturbations onto the footage, manifesting as jitter, inadvertent panning, and axis misalignment \cite{mingkhwan2017digital, aguilar2016real}. These spatiotemporal incongruities disrupt the optical flow coherence, infusing the sequences with undesirable optical phenomena like motion-induced blur, consecutive frame misregistration, and depth-induced parallax anomalies. Such aberrations invariably compromise the optical clarity and spatial resolution of the sequences. Figure \ref{fig:success_fail} depicts examples of successful and unsuccessful visualizations of various detection cases. The network's responses to these detection cases are influenced by the inherent challenges associated with differentiating between objects and background noise, and handling occlusions within the given data. For the false negative case, the pedestrian, due to their relatively smaller size, has been incorrectly classified as part of the background. This is a common challenge in object detection, especially in complex urban environments where numerous similar-sized non-target objects, such as lampposts or trash bins, are present. The detector's sensitivity towards small objects can be adversely affected by this, leading to the missed detection seen. In the false positive case, two vehicles are closely aligned and incorrectly detected as a single target. This issue arises when dealing with real-world scenarios where objects often overlap or occlude one another. With uncertain boundary information between closely located or overlapping objects, the network may perceive these objects as a single entity, leading to false positives.

\end{enumerate}

\section{Conclusion}
\label{sec:conclusion}
A neuromorphic quadrotor-based dataset for pedestrian/vehicle detection and localization in urban environments is presented. The dataset is fully annotated and is open sourced along with all accompanying Python code for the community. The dataset is used to train VGG, MobileNet, and DenseNet SNNs to evaluate object detection and localization tasks through a comprehensive ablation study. Ten frame-based single- and two-stage object detector DNNs were also used to evaluate the mAP of the dataset on object detection tasks. The experimental accuracy results of SNNs are on-par with results observed with matching networks trained on other large scale, contemporary datasets underscoring the reliability of the presented data. In light of this, the presented dataset accentuates the potential pitfalls arising from drone-related visual anomalies and such motion-induced blurring, jitter-related frame misalignments, and nuanced scale and perspective shifts. Concurrently, the onus on algorithmic methodologies is to proficiently handle the pronounced sparsity and refined temporal dynamics inherent to event-driven visual data. Future work will extend this study to perform instance and semantic segmentation on a wider variety of urban scenes using a combination of event and RGB cameras. 
\bibliography{refs}

\begin{thebibliography}{115}
\providecommand{\natexlab}[1]{#1}
\providecommand{\url}[1]{{#1}}
\providecommand{\urlprefix}{URL }
\providecommand{\doi}[1]{\url{https://doi.org/#1}}
\providecommand{\eprint}[2][]{\url{#2}}
 \bibcommenthead

\bibitem[{Aguilar and Angulo(2016)}]{aguilar2016real}
Aguilar WG, Angulo C (2016) Real-time model-based video stabilization for
  microaerial vehicles. Neural processing letters 43:459--477

\bibitem[{Amir et~al(2017)Amir, Taba, Berg, Melano, McKinstry, di~Nolfo, Nayak,
  Andreopoulos, Garreau, Mendoza, Kusnitz, DeBole, Esser, Delbr{\"u}ck,
  Flickner, and Modha}]{Amir2017}
Amir A, Taba B, Berg DJ, et~al (2017) A low power, fully event-based gesture
  recognition system. 2017 IEEE Conference on Computer Vision and Pattern
  Recognition (CVPR) pp 7388--7397

\bibitem[{Anderson and Gaston(2013)}]{anderson2013lightweight}
Anderson K, Gaston KJ (2013) Lightweight unmanned aerial vehicles will
  revolutionize spatial ecology. Frontiers in Ecology and the Environment
  11(3):138--146

\bibitem[{Ba et~al(2016)Ba, Kiros, and Hinton}]{ba2016layer}
Ba JL, Kiros JR, Hinton GE (2016) Layer normalization. arXiv preprint
  arXiv:160706450

\bibitem[{Baldi and Sadowski(2013)}]{baldi2013understanding}
Baldi P, Sadowski PJ (2013) Understanding dropout. Advances in neural
  information processing systems 26

\bibitem[{Bardow et~al(2016)Bardow, Davison, and
  Leutenegger}]{bardow2016simultaneous}
Bardow P, Davison AJ, Leutenegger S (2016) Simultaneous optical flow and
  intensity estimation from an event camera. In: Proceedings of the IEEE
  conference on computer vision and pattern recognition, pp 884--892

\bibitem[{Barnas et~al(2019)Barnas, Darby, Vandeberg, Rockwell, and
  Ellis-Felege}]{barnas2019comparison}
Barnas AF, Darby BJ, Vandeberg GS, et~al (2019) A comparison of drone imagery
  and ground-based methods for estimating the extent of habitat destruction by
  lesser snow geese (anser caerulescens caerulescens) in la p{\'e}rouse bay.
  PLoS One 14(8):e0217049

\bibitem[{Binas et~al(2017)Binas, Neil, Liu, and Delbruck}]{binas2017ddd17}
Binas J, Neil D, Liu SC, et~al (2017) Ddd17: End-to-end davis driving dataset.
  arXiv preprint arXiv:171101458

\bibitem[{Bjorck et~al(2018)Bjorck, Gomes, Selman, and
  Weinberger}]{bjorck2018understanding}
Bjorck N, Gomes CP, Selman B, et~al (2018) Understanding batch normalization.
  Advances in neural information processing systems 31

\bibitem[{Bock et~al(2020)Bock, Krajewski, Moers, Runde, Vater, and
  Eckstein}]{bock2020ind}
Bock J, Krajewski R, Moers T, et~al (2020) The ind dataset: A drone dataset of
  naturalistic road user trajectories at german intersections. In: 2020 IEEE
  Intelligent Vehicles Symposium (IV), IEEE, pp 1929--1934

\bibitem[{Braun et~al(2019)Braun, Krebs, Flohr, and
  Gavrila}]{braun2019eurocity}
Braun M, Krebs S, Flohr F, et~al (2019) Eurocity persons: A novel benchmark for
  person detection in traffic scenes. IEEE transactions on pattern analysis and
  machine intelligence 41(8):1844--1861

\bibitem[{Breuer et~al(2020)Breuer, Term{\"o}hlen, Homoceanu, and
  Fingscheidt}]{breuer2020opendd}
Breuer A, Term{\"o}hlen JA, Homoceanu S, et~al (2020) opendd: A large-scale
  roundabout drone dataset. In: 2020 IEEE 23rd International Conference on
  Intelligent Transportation Systems (ITSC), IEEE, pp 1--6

\bibitem[{Brutzkus and Globerson(2019)}]{brutzkus2019larger}
Brutzkus A, Globerson A (2019) Why do larger models generalize better? a
  theoretical perspective via the xor problem. In: International Conference on
  Machine Learning, PMLR, pp 822--830

\bibitem[{del Cerro et~al(2021)del Cerro, Cruz~Ulloa, Barrientos, and
  de~Le{\'o}n~Rivas}]{del2021unmanned}
del Cerro J, Cruz~Ulloa C, Barrientos A, et~al (2021) Unmanned aerial vehicles
  in agriculture: A survey. Agronomy 11(2):203

\bibitem[{Chen et~al(2012)Chen, Akselrod, Zhao, Perez~Carrasco,
  Linares-Barranco, and Culurciello}]{postures-dvs}
Chen S, Akselrod P, Zhao B, et~al (2012) Efficient feedforward categorization
  of objects and human postures with address-event image sensors. IEEE
  Transactions on Pattern Analysis and Machine Intelligence 34(2):302--314.
  \doi{10.1109/TPAMI.2011.120}

\bibitem[{Chen et~al(2018)Chen, Wang, Lu, Chen, and Wang}]{chen2018large}
Chen Y, Wang Y, Lu P, et~al (2018) Large-scale structure from motion with
  semantic constraints of aerial images. In: Chinese Conference on Pattern
  Recognition and Computer Vision (PRCV), Springer, pp 347--359

\bibitem[{Cordone et~al(2021)Cordone, Miramond, and
  Ferrante}]{cordone2021learning}
Cordone L, Miramond B, Ferrante S (2021) Learning from event cameras with
  sparse spiking convolutional neural networks. In: 2021 International Joint
  Conference on Neural Networks (IJCNN), IEEE, pp 1--8

\bibitem[{Cordone et~al(2022)Cordone, Miramond, and
  Thierion}]{Cordone_2022_IJCNN}
Cordone L, Miramond B, Thierion P (2022) Object detection with spiking neural
  networks on automotive event data. In: Proceedings of the IEEE International
  Joint Conference on Neural Networks (IJCNN)

\bibitem[{Daknama and Kraus(2017)}]{daknama2017vehicle}
Daknama R, Kraus E (2017) Vehicle routing with drones. arXiv preprint
  arXiv:170506431

\bibitem[{Dalal and Triggs(2005)}]{dalal2005histograms}
Dalal N, Triggs B (2005) Histograms of oriented gradients for human detection.
  In: 2005 IEEE computer society conference on computer vision and pattern
  recognition (CVPR'05), Ieee, pp 886--893

\bibitem[{De~Tournemire et~al(2020)De~Tournemire, Nitti, Perot, Migliore, and
  Sironi}]{de2020large}
De~Tournemire P, Nitti D, Perot E, et~al (2020) A large scale event-based
  detection dataset for automotive. arXiv preprint arXiv:200108499

\bibitem[{Dilshad et~al(2020)Dilshad, Hwang, Song, and
  Sung}]{dilshad2020applications}
Dilshad N, Hwang J, Song J, et~al (2020) Applications and challenges in video
  surveillance via drone: A brief survey. In: 2020 International Conference on
  Information and Communication Technology Convergence (ICTC), IEEE, pp
  728--732

\bibitem[{DJI(2022{\natexlab{a}})}]{DJIM100}
DJI (2022{\natexlab{a}}) Dji matrice m100: Quadcopter for developers.
  https://www.dji.com/matrice100

\bibitem[{DJI(2022{\natexlab{b}})}]{matricebattery}
DJI (2022{\natexlab{b}}) Matrice 100 tb48d battery.
  https://store.dji.com/product/matrice-100-tb48d-battery

\bibitem[{Doll{\'a}r et~al(2009)Doll{\'a}r, Wojek, Schiele, and
  Perona}]{dollar2009pedestrian}
Doll{\'a}r P, Wojek C, Schiele B, et~al (2009) Pedestrian detection: A
  benchmark. In: 2009 IEEE conference on computer vision and pattern
  recognition, IEEE, pp 304--311

\bibitem[{Dongare et~al(2012)Dongare, Kharde, Kachare
  et~al}]{dongare2012introduction}
Dongare A, Kharde R, Kachare AD, et~al (2012) Introduction to artificial neural
  network. International Journal of Engineering and Innovative Technology
  (IJEIT) 2(1):189--194

\bibitem[{Du et~al(2018)Du, Qi, Yu, Yang, Duan, Li, Zhang, Huang, and
  Tian}]{du2018unmanned}
Du D, Qi Y, Yu H, et~al (2018) The unmanned aerial vehicle benchmark: Object
  detection and tracking. In: Proceedings of the European conference on
  computer vision (ECCV), pp 370--386

\bibitem[{El~Shair and Rawashdeh(2022)}]{el2022high}
El~Shair Z, Rawashdeh SA (2022) High-temporal-resolution object detection and
  tracking using images and events. Journal of Imaging 8(8):210

\bibitem[{Enzweiler and Gavrila(2008)}]{enzweiler2008monocular}
Enzweiler M, Gavrila DM (2008) Monocular pedestrian detection: Survey and
  experiments. IEEE transactions on pattern analysis and machine intelligence
  31(12):2179--2195

\bibitem[{Escobar et~al(2009)Escobar, Masson, Vi{\'e}ville, and
  Kornprobst}]{Escobar2009ActionRU}
Escobar MJ, Masson GS, Vi{\'e}ville T, et~al (2009) Action recognition using a
  bio-inspired feedforward spiking network. International Journal of Computer
  Vision 82:284--301.
  \urlprefix\url{https://api.semanticscholar.org/CorpusID:441561}

\bibitem[{Ess et~al(2008)Ess, Leibe, Schindler, and Van~Gool}]{ess2008mobile}
Ess A, Leibe B, Schindler K, et~al (2008) A mobile vision system for robust
  multi-person tracking. In: 2008 IEEE Conference on Computer Vision and
  Pattern Recognition, IEEE, pp 1--8

\bibitem[{Fang et~al(2021)Fang, Yu, Chen, Masquelier, Huang, and
  Tian}]{fang2021incorporating}
Fang W, Yu Z, Chen Y, et~al (2021) Incorporating learnable membrane time
  constant to enhance learning of spiking neural networks. In: Proceedings of
  the IEEE/CVF International Conference on Computer Vision, pp 2661--2671

\bibitem[{Fei-Fei et~al(2004)Fei-Fei, Fergus, and Perona}]{Caltech101}
Fei-Fei L, Fergus R, Perona P (2004) Learning generative visual models from few
  training examples: An incremental bayesian approach tested on 101 object
  categories. Computer Vision and Pattern Recognition Workshop

\bibitem[{Fromm et~al(2019)Fromm, Schubert, Castilla, Linke, and
  McDermid}]{fromm2019automated}
Fromm M, Schubert M, Castilla G, et~al (2019) Automated detection of conifer
  seedlings in drone imagery using convolutional neural networks. Remote
  Sensing 11(21):2585

\bibitem[{Gallego et~al(2020)Gallego, Delbr{\"u}ck, Orchard, Bartolozzi, Taba,
  Censi, Leutenegger, Davison, Conradt, Daniilidis et~al}]{gallego2020event}
Gallego G, Delbr{\"u}ck T, Orchard G, et~al (2020) Event-based vision: A
  survey. IEEE transactions on pattern analysis and machine intelligence
  44(1):154--180

\bibitem[{Gehrig and Scaramuzza(2023)}]{gehrig2023recurrent}
Gehrig M, Scaramuzza D (2023) Recurrent vision transformers for object
  detection with event cameras. In: Proceedings of the IEEE/CVF conference on
  computer vision and pattern recognition, pp 13884--13893

\bibitem[{Geiger et~al(2013)Geiger, Lenz, Stiller, and
  Urtasun}]{geiger2013vision}
Geiger A, Lenz P, Stiller C, et~al (2013) Vision meets robotics: The kitti
  dataset. The International Journal of Robotics Research 32(11):1231--1237

\bibitem[{Geman et~al(1992)Geman, Bienenstock, and Doursat}]{geman1992neural}
Geman S, Bienenstock E, Doursat R (1992) Neural networks and the bias/variance
  dilemma. Neural computation 4(1):1--58

\bibitem[{Gholamalinezhad and Khosravi(2020)}]{gholamalinezhad2020pooling}
Gholamalinezhad H, Khosravi H (2020) Pooling methods in deep neural networks, a
  review. arXiv preprint arXiv:200907485

\bibitem[{Google(2023{\natexlab{a}})}]{colab}
Google (2023{\natexlab{a}}) Colaboratory: Frequently asked questions.
  \url{https://research.google.com/colaboratory/faq.html}

\bibitem[{Google(2023{\natexlab{b}})}]{gcp}
Google (2023{\natexlab{b}}) Google cloud compute platform.
  \url{https://console.cloud.google.com}

\bibitem[{Guerrero-G{\'o}mez-Olmedo et~al(2015)Guerrero-G{\'o}mez-Olmedo,
  Torre-Jim{\'e}nez, L{\'o}pez-Sastre, Maldonado-Basc{\'o}n, and
  Onoro-Rubio}]{guerrero2015extremely}
Guerrero-G{\'o}mez-Olmedo R, Torre-Jim{\'e}nez B, L{\'o}pez-Sastre R, et~al
  (2015) Extremely overlapping vehicle counting. In: Pattern Recognition and
  Image Analysis: 7th Iberian Conference, IbPRIA 2015, Santiago de Compostela,
  Spain, June 17-19, 2015, Proceedings 7, Springer, pp 423--431

\bibitem[{Gupta and Verma(2022)}]{gupta2022monitoring}
Gupta H, Verma OP (2022) Monitoring and surveillance of urban road traffic
  using low altitude drone images: a deep learning approach. Multimedia Tools
  and Applications pp 1--21

\bibitem[{Harris et~al(2020)Harris, Millman, van~der Walt, Gommers, Virtanen,
  Cournapeau, Wieser, Taylor, Berg, Smith, Kern, Picus, Hoyer, van Kerkwijk,
  Brett, Haldane, del R{\'{i}}o, Wiebe, Peterson, G{\'{e}}rard-Marchant,
  Sheppard, Reddy, Weckesser, Abbasi, Gohlke, and Oliphant}]{numpy}
Harris CR, Millman KJ, van~der Walt SJ, et~al (2020) Array programming with
  {NumPy}. Nature 585(7825):357--362. \doi{10.1038/s41586-020-2649-2},
  \urlprefix\url{https://doi.org/10.1038/s41586-020-2649-2}

\bibitem[{Hirano et~al(2006)Hirano, Garcia, Sukthankar, and
  Hoogs}]{hirano2006industry}
Hirano Y, Garcia C, Sukthankar R, et~al (2006) Industry and object recognition:
  Applications, applied research and challenges. In: Toward category-level
  object recognition. Springer, p 49--64

\bibitem[{Hossain and Lee(2019)}]{hossain2019deep}
Hossain S, Lee Dj (2019) Deep learning-based real-time multiple-object
  detection and tracking from aerial imagery via a flying robot with gpu-based
  embedded devices. Sensors 19(15):3371

\bibitem[{Howard et~al(2017)Howard, Zhu, Chen, Kalenichenko, Wang, Weyand,
  Andreetto, and Adam}]{howard2017mobilenets}
Howard AG, Zhu M, Chen B, et~al (2017) Mobilenets: Efficient convolutional
  neural networks for mobile vision applications. arXiv preprint
  arXiv:170404861

\bibitem[{Huang et~al(2017)Huang, Liu, Van Der~Maaten, and
  Weinberger}]{huang2017densely}
Huang G, Liu Z, Van Der~Maaten L, et~al (2017) Densely connected convolutional
  networks. In: Proceedings of the IEEE conference on computer vision and
  pattern recognition, pp 4700--4708

\bibitem[{Iaboni et~al(2021)Iaboni, Patel, Lobo, Choi, and
  Abichandani}]{iaboni2021event}
Iaboni C, Patel H, Lobo D, et~al (2021) Event camera based real-time detection
  and tracking of indoor ground robots. IEEE Access 9:166588--166602

\bibitem[{Iaboni et~al(2022)Iaboni, Lobo, Choi, and
  Abichandani}]{iaboni2022event}
Iaboni C, Lobo D, Choi JW, et~al (2022) Event-based motion capture system for
  online multi-quadrotor localization and tracking. Sensors 22(9):3240

\bibitem[{Insights(2024)}]{aerial-image-market}
Insights FB (2024) Aerial imaging market size, growth, analysis, report, 2032.
  https://bit.ly/3TE4Io2

\bibitem[{Ioffe and Szegedy(2015)}]{ioffe2015batch}
Ioffe S, Szegedy C (2015) Batch normalization: Accelerating deep network
  training by reducing internal covariate shift. In: International conference
  on machine learning, pmlr, pp 448--456

\bibitem[{Jocher et~al(2023)Jocher, Chaurasia, and
  Qiu}]{Jocher_YOLO_by_Ultralytics_2023}
Jocher G, Chaurasia A, Qiu J (2023) {YOLO by Ultralytics}.
  \urlprefix\url{https://github.com/ultralytics/ultralytics}

\bibitem[{Jocher et~al(2020)}]{glenn_jocher_2020_4154370}
Jocher G, et~al (2020) ultralytics/yolov5. \doi{10.5281/zenodo.4154370},
  \urlprefix\url{https://doi.org/10.5281/zenodo.4154370}

\bibitem[{Joyce et~al(2018)Joyce, Duce, Leahy, Leon, and
  Maier}]{joyce2018principles}
Joyce K, Duce S, Leahy S, et~al (2018) Principles and practice of acquiring
  drone-based image data in marine environments. Marine and Freshwater Research
  70(7):952--963

\bibitem[{Kim et~al(2019)Kim, Park, Na, and Yoon}]{kim2019spikingyolo}
Kim S, Park S, Na B, et~al (2019) Spiking-yolo: Spiking neural network for
  energy-efficient object detection. \eprint{1903.06530}

\bibitem[{Koch et~al(2016)Koch, Weber, Sobottka, Fladung, Clemens, and
  Berghold}]{koch2016outdoor}
Koch S, Weber T, Sobottka C, et~al (2016) Outdoor electroluminescence imaging
  of crystalline photovoltaic modules: Comparative study between manual
  ground-level inspections and drone-based aerial surveys. In: 32nd European
  Photovoltaic Solar Energy Conference and Exhibition, pp 1736--1740

\bibitem[{Kugele et~al(2020)Kugele, Pfeil, Pfeiffer, and
  Chicca}]{kugele2020efficient}
Kugele A, Pfeil T, Pfeiffer M, et~al (2020) Efficient processing of
  spatio-temporal data streams with spiking neural networks. Frontiers in
  Neuroscience 14:439

\bibitem[{Lawrence and Giles(2000)}]{lawrence2000overfitting}
Lawrence S, Giles CL (2000) Overfitting and neural networks: conjugate gradient
  and backpropagation. In: Proceedings of the IEEE-INNS-ENNS International
  Joint Conference on Neural Networks. IJCNN 2000. Neural Computing: New
  Challenges and Perspectives for the New Millennium, IEEE, pp 114--119

\bibitem[{Li et~al(2022)Li, Li, Jiang, Weng, Geng, Li, Ke, Li, Cheng, Nie
  et~al}]{li2022yolov6}
Li C, Li L, Jiang H, et~al (2022) Yolov6: A single-stage object detection
  framework for industrial applications. arXiv preprint arXiv:220902976

\bibitem[{Li et~al(2023)Li, Li, and Tian}]{li2023sodformer}
Li D, Li J, Tian Y (2023) Sodformer: Streaming object detection with
  transformer using events and frames. IEEE Transactions on Pattern Analysis
  and Machine Intelligence

\bibitem[{Li et~al(2017)Li, Liu, Ji, Li, and Shi}]{cifar10-dvs}
Li H, Liu H, Ji X, et~al (2017) Cifar10-dvs: An event-stream dataset for object
  classification. Frontiers in Neuroscience 11. \doi{10.3389/fnins.2017.00309},
  \urlprefix\url{https://www.frontiersin.org/article/10.3389/fnins.2017.00309}

\bibitem[{Li et~al(2019)Li, Dong, Yu, Tian, and Huang}]{li2019event}
Li J, Dong S, Yu Z, et~al (2019) Event-based vision enhanced: A joint detection
  framework in autonomous driving. In: 2019 ieee international conference on
  multimedia and expo (icme), IEEE, pp 1396--1401

\bibitem[{Liang and Delahaye(2019)}]{liang2019drone}
Liang M, Delahaye D (2019) Drone fleet deployment strategy for large scale
  agriculture and forestry surveying. In: 2019 IEEE Intelligent Transportation
  Systems Conference (ITSC), IEEE, pp 4495--4500

\bibitem[{Lin et~al(2014)Lin, Maire, Belongie, Hays, Perona, Ramanan,
  Doll{\'a}r, and Zitnick}]{lin2014microsoft}
Lin TY, Maire M, Belongie S, et~al (2014) Microsoft coco: Common objects in
  context. In: Computer Vision--ECCV 2014: 13th European Conference, Zurich,
  Switzerland, September 6-12, 2014, Proceedings, Part V 13, Springer, pp
  740--755

\bibitem[{Lin et~al(2017)Lin, Goyal, Girshick, He, and
  Doll{\'a}r}]{lin2017focal}
Lin TY, Goyal P, Girshick R, et~al (2017) Focal loss for dense object
  detection. In: Proceedings of the IEEE international conference on computer
  vision, pp 2980--2988

\bibitem[{Liu et~al(2016)Liu, Anguelov, Erhan, Szegedy, Reed, Fu, and
  Berg}]{liu2016ssd}
Liu W, Anguelov D, Erhan D, et~al (2016) Ssd: Single shot multibox detector.
  In: Computer Vision--ECCV 2016: 14th European Conference, Amsterdam, The
  Netherlands, October 11--14, 2016, Proceedings, Part I 14, Springer, pp
  21--37

\bibitem[{Loshchilov and Hutter(2019)}]{loshchilov2019decoupled}
Loshchilov I, Hutter F (2019) Decoupled weight decay regularization.
  \eprint{1711.05101}

\bibitem[{Marathe(2019)}]{marathe2019leveraging}
Marathe S (2019) Leveraging drone based imaging technology for pipeline and rou
  monitoring survey. In: SPE Asia Pacific Health, Safety, Security, Environment
  and Social Responsibility Symposium?, SPE, p D021S006R001

\bibitem[{Miao et~al(2019)Miao, Chen, Ning, Zi, Ren, Bing, and
  Knoll}]{miao2019neuromorphic}
Miao S, Chen G, Ning X, et~al (2019) Neuromorphic benchmark datasets for
  pedestrian detection, action recognition, and fall detection. Frontiers in
  neurorobotics 13:38

\bibitem[{Mingkhwan and Khawsuk(2017)}]{mingkhwan2017digital}
Mingkhwan E, Khawsuk W (2017) Digital image stabilization technique for fixed
  camera on small size drone. In: 2017 Third Asian Conference on Defence
  Technology (ACDT), IEEE, pp 12--19

\bibitem[{Mitrokhin et~al(2019)Mitrokhin, Ye, Ferm{\"u}ller, Aloimonos, and
  Delbruck}]{mitrokhin2019ev}
Mitrokhin A, Ye C, Ferm{\"u}ller C, et~al (2019) Ev-imo: Motion segmentation
  dataset and learning pipeline for event cameras. In: 2019 IEEE/RSJ
  International Conference on Intelligent Robots and Systems (IROS), IEEE, pp
  6105--6112

\bibitem[{Mittal et~al(2020)Mittal, Singh, and Sharma}]{mittal2020deep}
Mittal P, Singh R, Sharma A (2020) Deep learning-based object detection in
  low-altitude uav datasets: A survey. Image and Vision computing 104:104046

\bibitem[{Moeys et~al(2018)Moeys, Neil, Corradi, Kerr, Vance, Das, Coleman,
  McGinnity, Kerr, and Delbruck}]{pred18}
Moeys DP, Neil D, Corradi F, et~al (2018) Pred18: Dataset and further
  experiments with davis event camera in predator-prey robot chasing. IEEE
  Fourth International Conference on Event-Based Control, Communication and
  Signal Processing (EBCCSP) 2018 \doi{10.48550/ARXIV.1807.03128},
  \urlprefix\url{https://arxiv.org/abs/1807.03128}

\bibitem[{Mueller et~al(2016)Mueller, Smith, and Ghanem}]{mueller2016benchmark}
Mueller M, Smith N, Ghanem B (2016) A benchmark and simulator for uav tracking.
  In: Computer Vision--ECCV 2016: 14th European Conference, Amsterdam, The
  Netherlands, October 11--14, 2016, Proceedings, Part I 14, Springer, pp
  445--461

\bibitem[{Nagi et~al(2011)Nagi, Ducatelle, Di~Caro, Cire{\c{s}}an, Meier,
  Giusti, Nagi, Schmidhuber, and Gambardella}]{nagi2011max}
Nagi J, Ducatelle F, Di~Caro GA, et~al (2011) Max-pooling convolutional neural
  networks for vision-based hand gesture recognition. In: 2011 IEEE
  international conference on signal and image processing applications
  (ICSIPA), IEEE, pp 342--347

\bibitem[{Negri et~al(2018{\natexlab{a}})Negri, Soto, Linares-Barranco, and
  Serrano-Gotarredona}]{scenes-dvs}
Negri P, Soto M, Linares-Barranco B, et~al (2018{\natexlab{a}}) Scene context
  classification with event-driven spiking deep neural networks. In: 2018 25th
  IEEE International Conference on Electronics, Circuits and Systems (ICECS),
  pp 569--572, \doi{10.1109/ICECS.2018.8617982}

\bibitem[{Negri et~al(2018{\natexlab{b}})Negri, Soto, Linares-Barranco, and
  Serrano-Gotarredona}]{negri2018scene}
Negri P, Soto M, Linares-Barranco B, et~al (2018{\natexlab{b}}) Scene context
  classification with event-driven spiking deep neural networks. In: 2018 25th
  IEEE International Conference on Electronics, Circuits and Systems (ICECS),
  IEEE, pp 569--572

\bibitem[{Nvidia(2022)}]{xavier}
Nvidia (2022) Nvidia jetson agx xavier development kit.
  https://www.nvidia.com/en-us/autonomous-machines/embedded-systems/jetson-agx-xavier/

\bibitem[{Orchard et~al(2015)Orchard, Jayawant, Cohen, and Thakor}]{nmnist}
Orchard G, Jayawant A, Cohen GK, et~al (2015) Converting static image datasets
  to spiking neuromorphic datasets using saccades. Frontiers in Neuroscience 9.
  \doi{10.3389/fnins.2015.00437},
  \urlprefix\url{https://www.frontiersin.org/article/10.3389/fnins.2015.00437}

\bibitem[{Orchard et~al(2021)Orchard, Frady, Rubin, Sanborn, Shrestha, Sommer,
  and Davies}]{orchard2021efficient}
Orchard G, Frady EP, Rubin DBD, et~al (2021) Efficient neuromorphic signal
  processing with loihi 2. In: 2021 IEEE Workshop on Signal Processing Systems
  (SiPS), IEEE, pp 254--259

\bibitem[{Pang et~al(2020)Pang, Cao, Li, Xie, Sun, and Gong}]{pang2020tju}
Pang Y, Cao J, Li Y, et~al (2020) Tju-dhd: A diverse high-resolution dataset
  for object detection. IEEE Transactions on Image Processing 30:207--219

\bibitem[{Papadopoulos et~al(2017)Papadopoulos, Uijlings, Keller, and
  Ferrari}]{papadopoulos2017}
Papadopoulos DP, Uijlings JR, Keller F, et~al (2017) Extreme clicking for
  efficient object annotation. In: Proceedings of the IEEE international
  conference on computer vision, pp 4930--4939

\bibitem[{Perot et~al(2020)Perot, De~Tournemire, Nitti, Masci, and
  Sironi}]{perot2020learning}
Perot E, De~Tournemire P, Nitti D, et~al (2020) Learning to detect objects with
  a 1 megapixel event camera. Advances in Neural Information Processing Systems
  33:16639--16652

\bibitem[{Prophesee(2024)}]{prophesee-params}
Prophesee (2024) Gen3.1 vga sensor. https://bit.ly/4cGe2jK

\bibitem[{Puri(2005)}]{puri2005survey}
Puri A (2005) A survey of unmanned aerial vehicles (uav) for traffic
  surveillance. Department of computer science and engineering, University of
  South Florida pp 1--29

\bibitem[{Pérez-Carrasco et~al(2013)Pérez-Carrasco, Zhao, Serrano, Acha,
  Serrano-Gotarredona, Chen, and Linares-Barranco}]{poker-dvs}
Pérez-Carrasco J, Zhao B, Serrano C, et~al (2013) Mapping from frame-driven to
  frame-free event-driven vision systems by low-rate rate-coding and
  coincidence processing. application to feed forward convnets. IEEE
  Transactions on Pattern Analysis and Machine Intelligence 35:2706 -- 2719.
  \doi{10.1109/TPAMI.2013.71}

\bibitem[{Ramachandran and Sangaiah(2021)}]{ramachandran2021review}
Ramachandran A, Sangaiah AK (2021) A review on object detection in unmanned
  aerial vehicle surveillance. International Journal of Cognitive Computing in
  Engineering 2:215--228

\bibitem[{Rebecq et~al(2018)Rebecq, Gallego, Mueggler, and
  Scaramuzza}]{rebecq2018reconstruction}
Rebecq H, Gallego G, Mueggler E, et~al (2018) Emvs: Event-based multi-view
  stereo—3d reconstruction with an event camera in real-time. International
  Journal of Computer Vision 126. \doi{10.1007/s11263-017-1050-6}

\bibitem[{Reckling et~al(2021)Reckling, Mitasova, Wegmann, Kauffman, and
  Reid}]{reckling2021efficient}
Reckling W, Mitasova H, Wegmann K, et~al (2021) Efficient drone-based rare
  plant monitoring using a species distribution model and ai-based object
  detection. Drones 5(4):110

\bibitem[{Ren et~al(2015)Ren, He, Girshick, and Sun}]{ren2015faster}
Ren S, He K, Girshick R, et~al (2015) Faster r-cnn: Towards real-time object
  detection with region proposal networks. Advances in neural information
  processing systems 28

\bibitem[{Rold{\'a}n-G{\'o}mez et~al(2021)Rold{\'a}n-G{\'o}mez,
  Gonz{\'a}lez-Gironda, and Barrientos}]{roldan2021survey}
Rold{\'a}n-G{\'o}mez JJ, Gonz{\'a}lez-Gironda E, Barrientos A (2021) A survey
  on robotic technologies for forest firefighting: Applying drone swarms to
  improve firefighters’ efficiency and safety. Applied Sciences 11(1):363

\bibitem[{Russakovsky et~al(2015)Russakovsky, Deng, Su, Krause, Satheesh, Ma,
  Huang, Karpathy, Khosla, Bernstein et~al}]{russakovsky2015}
Russakovsky O, Deng J, Su H, et~al (2015) Imagenet large scale visual
  recognition challenge. International journal of computer vision 115:211--252

\bibitem[{Sandino et~al(2020)Sandino, Vanegas, Maire, Caccetta, Sanderson, and
  Gonzalez}]{sandino2020uav}
Sandino J, Vanegas F, Maire F, et~al (2020) Uav framework for autonomous
  onboard navigation and people/object detection in cluttered indoor
  environments. Remote Sensing 12(20):3386

\bibitem[{Sandino et~al(2021)Sandino, Maire, Caccetta, Sanderson, and
  Gonzalez}]{sandino2021drone}
Sandino J, Maire F, Caccetta P, et~al (2021) Drone-based autonomous motion
  planning system for outdoor environments under object detection uncertainty.
  Remote Sensing 13(21):4481

\bibitem[{Santurkar et~al(2018)Santurkar, Tsipras, Ilyas, and
  Madry}]{santurkar2018does}
Santurkar S, Tsipras D, Ilyas A, et~al (2018) How does batch normalization help
  optimization? Advances in neural information processing systems 31

\bibitem[{Scherer et~al(2010)Scherer, M{\"u}ller, and
  Behnke}]{scherer2010evaluation}
Scherer D, M{\"u}ller A, Behnke S (2010) Evaluation of pooling operations in
  convolutional architectures for object recognition. In: International
  conference on artificial neural networks, Springer, pp 92--101

\bibitem[{Seifert et~al(2019)Seifert, Seifert, Vogt, Drew, Van~Aardt, Kunneke,
  and Seifert}]{seifert2019influence}
Seifert E, Seifert S, Vogt H, et~al (2019) Influence of drone altitude, image
  overlap, and optical sensor resolution on multi-view reconstruction of forest
  images. Remote sensing 11(10):1252

\bibitem[{Serrano-Gotarredona and Linares-Barranco(2013)}]{mnist-dvs}
Serrano-Gotarredona T, Linares-Barranco B (2013) A 128x128 1.5\% contrast
  sensitivity 0.9\% fpn 3 µs latency 4 mw asynchronous frame-free dynamic
  vision sensor using transimpedance preamplifiers. IEEE Journal of Solid-State
  Circuits 48(3):827--838. \doi{10.1109/JSSC.2012.2230553}

\bibitem[{Shazeer et~al(2017)Shazeer, Mirhoseini, Maziarz, Davis, Le, Hinton,
  and Dean}]{shazeer2017outrageously}
Shazeer N, Mirhoseini A, Maziarz K, et~al (2017) Outrageously large neural
  networks: The sparsely-gated mixture-of-experts layer. arXiv preprint
  arXiv:170106538

\bibitem[{Shi et~al(2015)Shi, Chen, Wang, Yeung, Wong, and
  Woo}]{shi2015convolutional}
Shi X, Chen Z, Wang H, et~al (2015) Convolutional lstm network: A machine
  learning approach for precipitation nowcasting. Advances in neural
  information processing systems 28

\bibitem[{Simonyan and Zisserman(2014)}]{simonyan2014very}
Simonyan K, Zisserman A (2014) Very deep convolutional networks for large-scale
  image recognition. arXiv preprint arXiv:14091556

\bibitem[{Sironi et~al(2018)Sironi, Brambilla, Bourdis, Lagorce, and
  Benosman}]{sironi2018hats}
Sironi A, Brambilla M, Bourdis N, et~al (2018) Hats: Histograms of averaged
  time surfaces for robust event-based object classification. In: Proceedings
  of the IEEE conference on computer vision and pattern recognition, pp
  1731--1740

\bibitem[{Srivastava et~al(2014)Srivastava, Hinton, Krizhevsky, Sutskever, and
  Salakhutdinov}]{srivastava2014dropout}
Srivastava N, Hinton G, Krizhevsky A, et~al (2014) Dropout: a simple way to
  prevent neural networks from overfitting. The journal of machine learning
  research 15(1):1929--1958

\bibitem[{Su et~al(2012)Su, Deng, and Fei-Fei}]{su2012}
Su H, Deng J, Fei-Fei L (2012) Crowdsourcing annotations for visual object
  detection. In: Workshops at the twenty-sixth AAAI conference on artificial
  intelligence, Citeseer

\bibitem[{Sun et~al(2017)Sun, Song, Jiang, Pan, and Pang}]{sun2017learning}
Sun M, Song Z, Jiang X, et~al (2017) Learning pooling for convolutional neural
  network. Neurocomputing 224:96--104

\bibitem[{Varghese et~al(2017)Varghese, Gubbi, Sharma, and
  Balamuralidhar}]{varghese2017power}
Varghese A, Gubbi J, Sharma H, et~al (2017) Power infrastructure monitoring and
  damage detection using drone captured images. In: 2017 international joint
  conference on neural networks (IJCNN), IEEE, pp 1681--1687

\bibitem[{Vasudevan et~al(2020)Vasudevan, Negri, Linares-Barranco, and
  Serrano-Gotarredona}]{sl-animals-dvs}
Vasudevan A, Negri P, Linares-Barranco B, et~al (2020) Introduction and
  analysis of an event-based sign language dataset. In: 2020 15th IEEE
  International Conference on Automatic Face and Gesture Recognition (FG 2020),
  \doi{10.1109/FG47880.2020.00069}

\bibitem[{Wang et~al(2022)Wang, Bochkovskiy, and Liao}]{wang2022yolov7}
Wang CY, Bochkovskiy A, Liao HYM (2022) Yolov7: Trainable bag-of-freebies sets
  new state-of-the-art for real-time object detectors. arXiv preprint
  arXiv:220702696

\bibitem[{Wen et~al(2020)Wen, Du, Cai, Lei, Chang, Qi, Lim, Yang, and
  Lyu}]{wen2020ua}
Wen L, Du D, Cai Z, et~al (2020) Ua-detrac: A new benchmark and protocol for
  multi-object detection and tracking. Computer Vision and Image Understanding
  193:102907

\bibitem[{Wojek et~al(2009)Wojek, Walk, and Schiele}]{wojek2009multi}
Wojek C, Walk S, Schiele B (2009) Multi-cue onboard pedestrian detection. In:
  2009 IEEE Conference on Computer Vision and Pattern Recognition, IEEE, pp
  794--801

\bibitem[{Zhang et~al(2017)Zhang, Benenson, and Schiele}]{zhang2017citypersons}
Zhang S, Benenson R, Schiele B (2017) Citypersons: A diverse dataset for
  pedestrian detection. In: Proceedings of the IEEE conference on computer
  vision and pattern recognition, pp 3213--3221

\bibitem[{Zheng et~al(2017)Zheng, Zhang, Sun, Chandraker, Yang, and
  Tian}]{zheng2017person}
Zheng L, Zhang H, Sun S, et~al (2017) Person re-identification in the wild. In:
  Proceedings of the IEEE conference on computer vision and pattern
  recognition, pp 1367--1376

\bibitem[{Zhilenkov and Epifantsev(2018)}]{zhilenkov2018system}
Zhilenkov AA, Epifantsev IR (2018) System of autonomous navigation of the drone
  in difficult conditions of the forest trails. In: 2018 IEEE Conference of
  Russian Young Researchers in Electrical and Electronic Engineering
  (EIConRus), IEEE, pp 1036--1039

\bibitem[{Zhu et~al(2021)Zhu, Wen, Du, Bian, Fan, Hu, and Ling}]{vizdrone}
Zhu P, Wen L, Du D, et~al (2021) Detection and tracking meet drones challenge.
  IEEE Transactions on Pattern Analysis and Machine Intelligence pp 1--1.
  \doi{10.1109/TPAMI.2021.3119563}

\end{thebibliography}

\end{document}